\documentclass[letterpaper]{article}
\usepackage[numbers]{natbib}
\usepackage[preprint]{aaai2027_custom}
\usepackage{graphicx}
\usepackage{amsmath}
\usepackage{amssymb}
\usepackage{url}
\usepackage{booktabs}
\usepackage{dblfloatfix}
\usepackage{tocloft}
\usepackage[colorlinks=true,
            linkcolor=red,
            citecolor=red,
            urlcolor=red]{hyperref}
\title{How to Tame Grokking: Representation Geometry as a Control Signal}
\author{
    Maksim A. Kazanskii
}

\affiliations{
    Independent Researcher\\
    mkazanskii@gmail.com
}
\begin{document}
\maketitle
\thispagestyle{plain}
\pagestyle{plain}

\begin{abstract}
Grokking is a phenomenon in which neural networks initially memorize training data and only later exhibit strong generalization after prolonged optimization. Despite extensive recent study, the factors influencing the emergence and timing of grokking remain incompletely understood. We investigate the relationship between representation geometry and delayed generalization. We find that dimensionality collapse consistently precedes the onset of grokking in all evaluated settings. Motivated by these observations, we introduce Geometric Dimensionality Regularization (GeomDR), a simple spectral regularizer that modifies the effective dimensionality of hidden representations during training. Across modular addition, modular division, and permutation composition tasks, GeomDR consistently alters grokking dynamics and can substantially accelerate the onset of generalization depending on the intervention schedule and target dimensionality. In several settings, grokking is accelerated by up to $52$ times relative to standard AdamW training. Similar qualitative effects are observed in both multilayer perceptrons and transformers. Together, these results suggest that representation geometry can serve as an effective control signal for grokking and provide evidence that geometric interventions offer a practical approach for studying and influencing delayed generalization in neural networks.
\end{abstract}
\section{Introduction}

Grokking is a delayed-generalization phenomenon in which neural networks first memorize training data and only much later achieve strong test performance after prolonged optimization \cite{power2022grokking}. This behavior differs from conventional learning dynamics, where training and test performance typically improve together, and has become a useful setting for studying generalization in overparameterized neural networks \cite{power2022grokking,liu2022towards,nanda2023progress,varma2023grokking}.

Prior work has linked grokking to weight decay, feature compression, circuit formation, and representation learning dynamics \cite{power2022grokking,nanda2023progress,liu2022towards}. A common theme in these explanations is that learned representations often become progressively compressed into lower-dimensional structures. More broadly, representation geometry and dimensionality have been shown to play important roles in optimization and generalization \cite{ansuini2019intrinsic,li2018measuring}.

Recent work has shown that grokking can be accelerated or altered through optimization dynamics, weight-norm control, sparse subnetworks, or embedding transfer \cite{liu2022omnigrok,minegishi2023bridging,lee2024grokfast,xu2025groktransfer}. However, comparatively less attention has been paid to controlling grokking through direct interventions on the geometry of hidden representations. If delayed generalization is closely connected to representation geometry, then explicitly modifying this geometry may influence the onset and speed of grokking.

We investigate this hypothesis by introducing Geometric Dimensionality Regularization (GeomDR), a representation-level spectral regularizer that suppresses variance outside a target subspace and thereby controls the effective dimensionality of hidden representations during training. GeomDR directly modifies the covariance spectrum of hidden activations.

We perform a systematic study across grokking tasks, architectures, intervention schedules, target dimensionalities, and random seeds. Our results show that geometric interventions can substantially alter delayed generalization dynamics, accelerating grokking by up to $52$ times in some settings and, under stronger interventions, delaying or suppressing generalization. We further find that changes in effective dimensionality consistently precede the transition from memorization to generalization, suggesting that representation dimensionality is not merely a diagnostic statistic but a controllable variable associated with grokking dynamics.

Our contributions are threefold:

\begin{itemize}
\item We show that direct geometric interventions can substantially alter delayed generalization, enabling acceleration and, in some settings, delay or suppression.
\item We introduce GeomDR, a representation-level spectral regularizer for directly controlling the effective dimensionality of hidden representations.
\item We provide a systematic empirical study of how intervention strength, timing, target dimensionality, task structure, and architecture affect grokking dynamics.

\end{itemize}

\section{Related Work}

Early work on grokking identified delayed generalization in algorithmic tasks and highlighted the importance of regularization, particularly weight decay, in the transition from memorization to generalization \cite{power2022grokking}. Subsequent studies connected grokking to feature compression, circuit formation, and representation learning dynamics \cite{liu2022towards,nanda2023progress,varma2023grokking}. These works largely treat representation geometry as an emergent property of learning, whereas our goal is to investigate whether geometry can be directly manipulated to control grokking dynamics.

Several approaches have sought to accelerate or modify grokking through changes to the training process. Omnigrok studies the role of weight norms and conventional regularization \cite{liu2022omnigrok}, Grokking Tickets relates delayed generalization to sparse subnetworks and pruning \cite{minegishi2023bridging}, Grokfast accelerates grokking through gradient filtering \cite{lee2024grokfast}, and GrokTransfer explores embedding transfer from weaker models \cite{xu2025groktransfer}. In contrast, GeomDR acts directly on hidden representations by modifying their covariance spectrum rather than altering optimization dynamics, sparsity, or transferred embeddings.

Representation geometry provides a useful framework for understanding learning dynamics. Prior work has shown that independently trained networks often converge to similar representational structures \cite{kornblith2019similarity}, while geometric analyses have been used to study feature organization and learning across architectures \cite{raghu2021vision}. Neural representations frequently exhibit low intrinsic dimensionality \cite{ansuini2019intrinsic,li2018measuring,aghajanyan2021intrinsic} and may undergo phases of compression and expansion during learning \cite{recanatesi2019compression}. Related evidence from Neural Collapse, Information Bottleneck analyses, and self-supervised learning further suggests that covariance structure and representation dimensionality are closely linked to learning and generalization \cite{papyan2020prevalence,han2022neuralcollapse,kothapalli2022review,tishby2015deep,shwartz2017opening,bardes2022vicreg,jing2022understanding}.

Existing grokking studies either analyze representation geometry as a correlate of delayed generalization or influence grokking indirectly through regularization, optimization, sparsity, or transfer mechanisms \cite{power2022grokking,liu2022towards,nanda2023progress,lee2024grokfast,xu2025groktransfer}. GeomDR instead treats representation geometry as the object of intervention itself, enabling controlled experiments on how effective dimensionality influences the onset and speed of grokking.

\section{Methods}

 We introduce a geometry-based regularization framework that directly modifies the effective dimensionality of hidden representations during training. 

We consider supervised algorithmic learning tasks with inputs $x$ and labels $y$. A neural network $f_{\theta}$ is trained to predict

\begin{equation}
\hat{y} = f_{\theta}(x).
\end{equation}

For each hidden layer $\ell$, the network produces a representation matrix

\begin{equation}
Z^{(\ell)} \in \mathbb{R}^{N \times d},
\end{equation}

where $N$ is the number of examples and $d$ is the representation dimension.

\subsection{Geometric Characterization of Representations}

For each representation matrix $Z$, we first center the representations:

\begin{equation}
\widetilde{Z} = Z - \bar{Z},
\end{equation}

where $\bar{Z}$ denotes the feature-wise mean of $Z$.

We then compute the empirical covariance matrix

\begin{equation}
C
=
\frac{1}{N-1}
\widetilde{Z}^{\top}\widetilde{Z}.
\end{equation}

Let $\mu_1,\ldots,\mu_d$ denote the eigenvalues of $C$.
Building on prior work on intrinsic dimensionality and representation
geometry \cite{ansuini2019intrinsic}, we compute \textbf{effective dimensionality} using the participation ratio

\begin{equation}
D_{\mathrm{eff}}
=
\frac{\left(\sum_i \mu_i\right)^2}
{\sum_i \mu_i^2}.
\end{equation}
Lower values of $D_{\mathrm{eff}}$ correspond to representations
whose variance is concentrated in a smaller number of directions.

Let each representation vector be normalized as

\begin{equation}
\hat{z}_i = \frac{z_i}{\|z_i\|_2}.
\end{equation}

Then we define \textbf{local neighborhood distance} as the average Euclidean distance to the
$k$ nearest neighbors of each normalized representation,

\begin{equation}
\rho
=
\frac{1}{N}
\sum_{i=1}^{N}
\frac{1}{k}
\sum_{j \in \mathcal{N}_k(i)}
\|\hat z_i-\hat z_j\|_2,
\end{equation}

where $\mathcal{N}_k(i)$ denotes the set of $k$ nearest neighbors of representation $i$. Unless otherwise specified, we use $k=10$ throughout all experiments.

\subsection{Geometric Dimensionality Regularization}
We hypothesize that the reduction of representation dimensionality is not merely a consequence of grokking but a driving factor in the transition from memorization to generalization. 
If representation dimensionality is mechanistically linked to grokking, then directly controlling the geometry of hidden representations should alter the timing and dynamics of the memorization-to-generalization transition.
Motivated by this idea, we \textbf{introduce Geometric Dimensionality
Regularization (GeomDR)}. For each hidden layer $\ell$, we compute the empirical covariance matrix

\begin{equation}
C^{(\ell)}
=
\frac{1}{N-1}
\left(\widetilde{Z}^{(\ell)}\right)^{\top}
\widetilde{Z}^{(\ell)}.
\end{equation}

Let

\begin{equation}
\mu_1^{(\ell)}
\ge
\mu_2^{(\ell)}
\ge
\cdots
\ge
\mu_d^{(\ell)}
\end{equation}

denote the eigenvalues of $C^{(\ell)}$ sorted in descending order.

Given a target dimensionality $d^*$, we define the layer-wise geometric regularization loss as

\begin{equation}
L_{\mathrm{GeomDR}}^{(\ell)}
=
\sum_{j=d^*+1}^{d}
\mu_j^{(\ell)}.
\end{equation}

This objective admits several geometric interpretations related
to effective-rank control, low rank covariance approximation,
and representation volume compression (see Appendix~\ref{appendix:spectral}).

Unless otherwise specified, GeomDR is applied to all hidden layers and excluded from the input embedding and output layers. This objective penalizes variance contained in directions beyond the leading $d^*$ principal components, encouraging representations to concentrate within a lower-dimensional subspace. The objective is therefore closely related to classical low rank approximation and principal component analysis \cite{eckart1936approximation,jolliffe2002principal}.

The total geometric regularization loss is

\begin{equation}
L_{\mathrm{GeomDR}}
=
\sum_{\ell}
L_{\mathrm{GeomDR}}^{(\ell)},
\end{equation}

The full training objective is

\begin{equation}
L =
L_{\mathrm{task}} + \lambda(t) L_{\mathrm{GeomDR}},
\end{equation}

where $L_{\mathrm{task}}$ denotes the task-specific training loss.

The geometric regularizer is activated after an initial training phase, allowing the network to fit the training data before constraining representation geometry. This design is motivated by grokking, where generalization typically emerges only after memorization \cite{power2022grokking}. To avoid an abrupt change in the optimization objective,
the regularization strength is introduced gradually using a
cosine ramp. Smooth scheduling of optimization hyperparameters
has become a common strategy in deep learning to improve
training stability and optimization behavior
\cite{loshchilov2017sgdr,goyal2017accurate}.

\begin{equation}
\lambda(t)=
\lambda_{\max}
\frac{1-\cos(\pi q(t))}{2},
\end{equation}
where

\begin{equation}
q(t)=
\begin{cases}
0,
& t < t_s, \\
\dfrac{t-t_s}{T_{\mathrm{ramp}}},
& t_s \le t < t_s+T_{\mathrm{ramp}}, \\
1,
& t \ge t_s+T_{\mathrm{ramp}}.
\end{cases}
\label{eq:ramp}
\end{equation}

Here $t_s$ is the activation time step, $T_{\mathrm{ramp}}$ is the ramp
duration, and $\lambda_{\max}$ is the final regularization strength.
For brevity, we refer to $\lambda_{\max}$ simply as $\lambda$ when
reporting experimental configurations.

GeomDR requires covariance estimation and eigendecomposition of hidden representations, resulting in a per-layer complexity of $O(Nd^2+d^3)$ (see Appendix~\ref{app:scalability}).

\subsection{Experimental Protocol}

We evaluate the proposed framework across three algorithmic learning tasks: modular addition, modular division, and permutation composition. These tasks are commonly used in the grokking literature because they exhibit delayed generalization under appropriate training conditions \cite{power2022grokking}. Experiments are conducted using two neural architectures: a multilayer perceptron (MLP) and a Transformer \cite{vaswani2017attention}. For all tasks, a fixed $30\%/70\%$ train--test split is used. Following prior grokking work \cite{power2022grokking}, no separate validation set is employed, and test-set dynamics are reported directly. Unless otherwise specified, all reported results are averaged across five independent random seeds. A run is considered to have successfully grokked when the training accuracy is $1.0$ and the test accuracy satisfies
\[
\mathrm{Acc}_{\mathrm{test}} \ge 0.999\,\mathrm{Acc}_{\mathrm{train}}
\]
for $10$ consecutive evaluations. The evaluation is performed every 10 optimization steps. Therefore this criterion corresponds to maintaining the target accuracy for 100 consecutive optimization steps. The grokking step is defined as the first optimization step at which this condition is satisfied.

For all tasks and architectures, geometric regularization is not applied at the beginning of training. Instead, the regularizer is activated after an initial unconstrained optimization phase. Unless otherwise specified, MLP experiments use an intervention activation step of $t_s = 2000$. Transformer experiments use $t_s = 1000$.
For each task--architecture pair, we use the unregularized model as the baseline reference. Our main experimental procedure consists of three stages:

\begin{enumerate}
\item \textbf{Schedule sweep (MLP).} For MLPs, we sweep both the final regularization strength
$\lambda_{\max}$ and ramp duration $T_{\mathrm{ramp}}$.
For Transformers, we sweep only $\lambda_{\max}$. Each configuration is evaluated across five random seeds and compared to the baseline. 

\item \textbf{Dimensionality sweep.} We perform a sweep over the target dimensionality $d^*$. 

\item \textbf{Intervention-start sweep.} We vary the activation step $t_s$ at which geometric regularization is introduced.

\end{enumerate}

Complete architectural specifications,
hyperparameter grids, and implementation details
are provided in Appendix~\ref{app:tasks_architectures}.

\section{Results}

\begin{figure*}[t]
\centering

\includegraphics[width=0.32\textwidth]{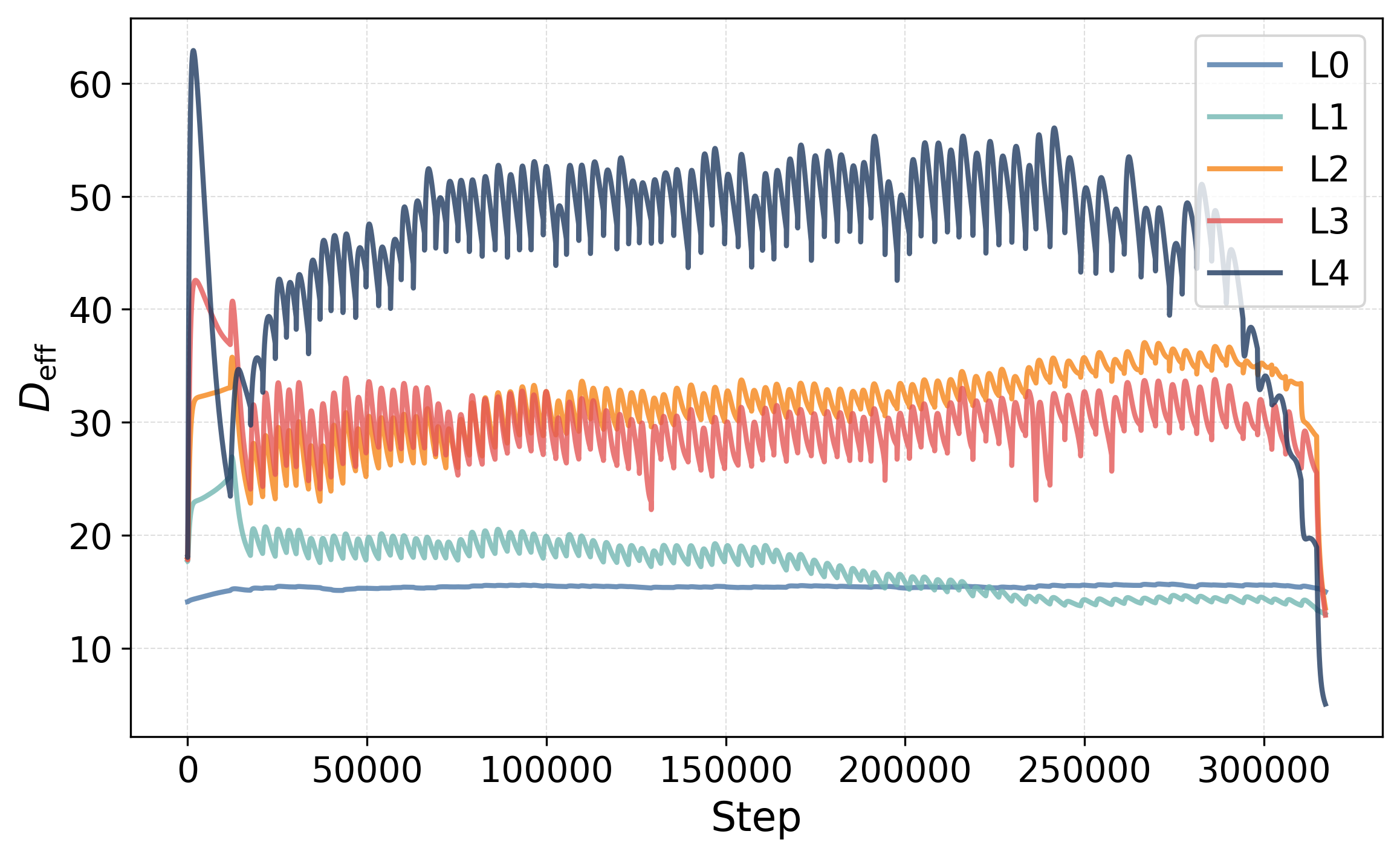}
\hfill
\includegraphics[width=0.32\textwidth]{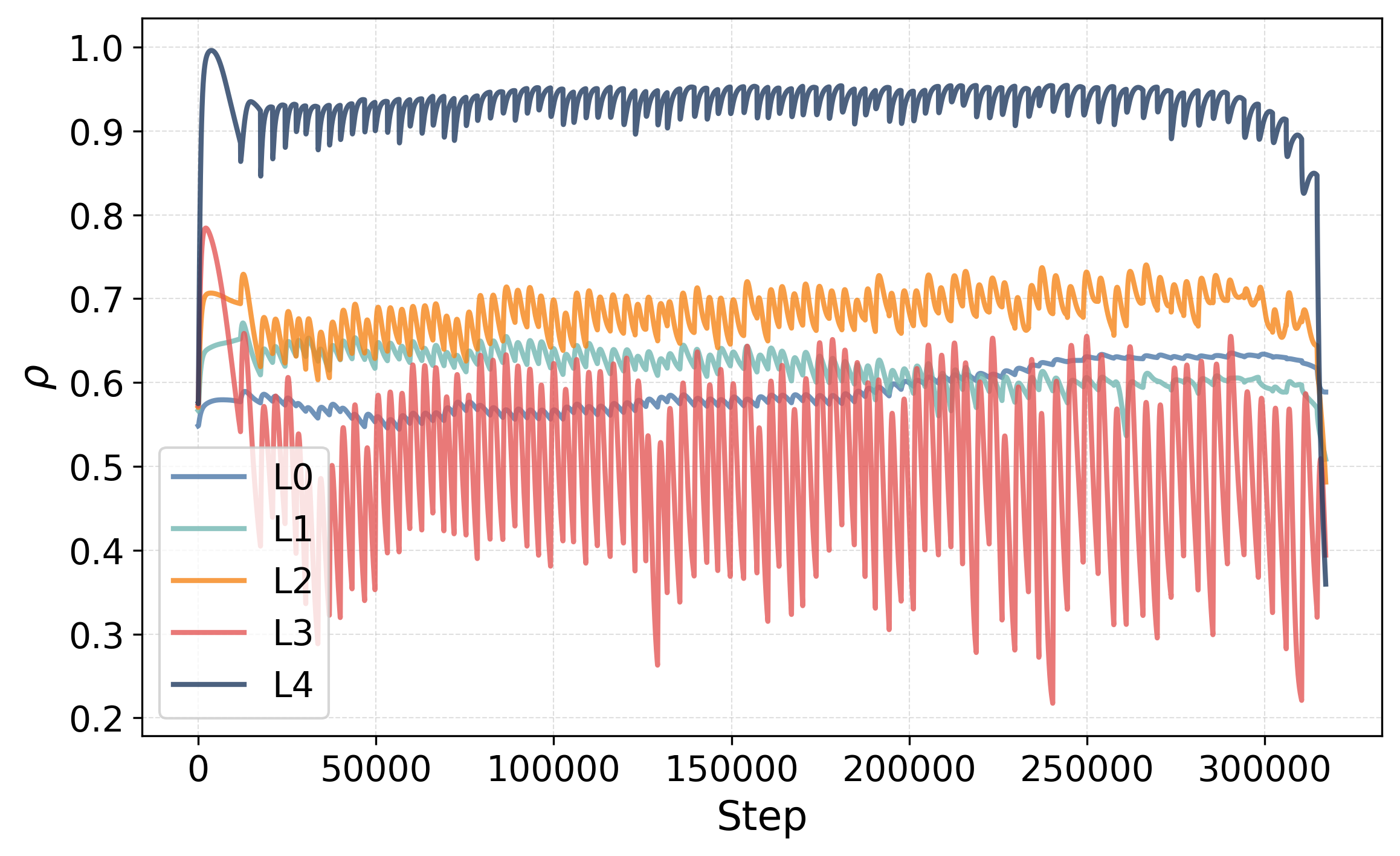}
\hfill
\includegraphics[width=0.32\textwidth]{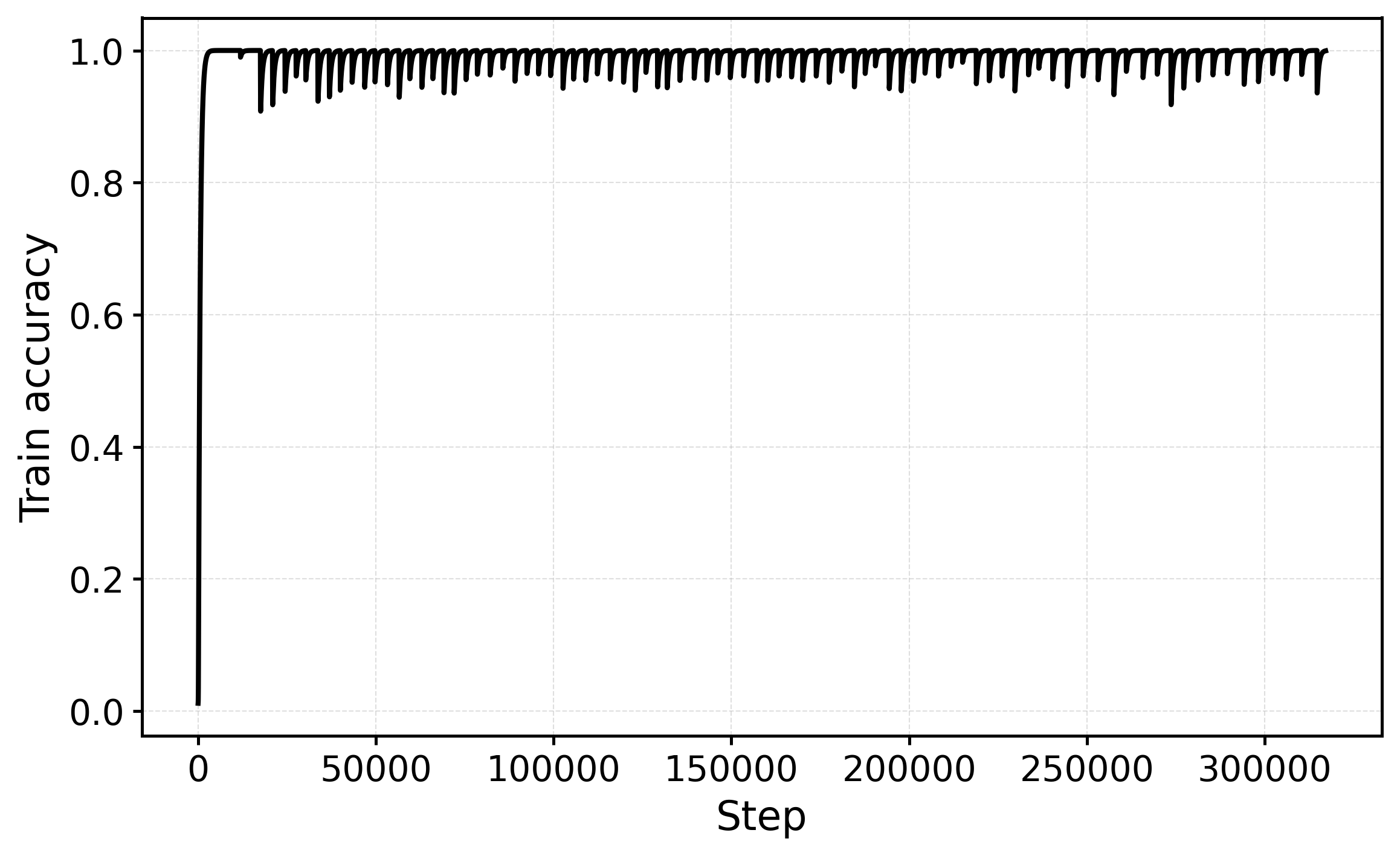}

\vspace{1mm}

{\small
(a) All steps: $D_{\mathrm{eff}}$
\hfill
(b) All steps: $\rho$
\hfill
(c) Train accuracy
}

\vspace{2mm}

\includegraphics[width=0.32\textwidth]{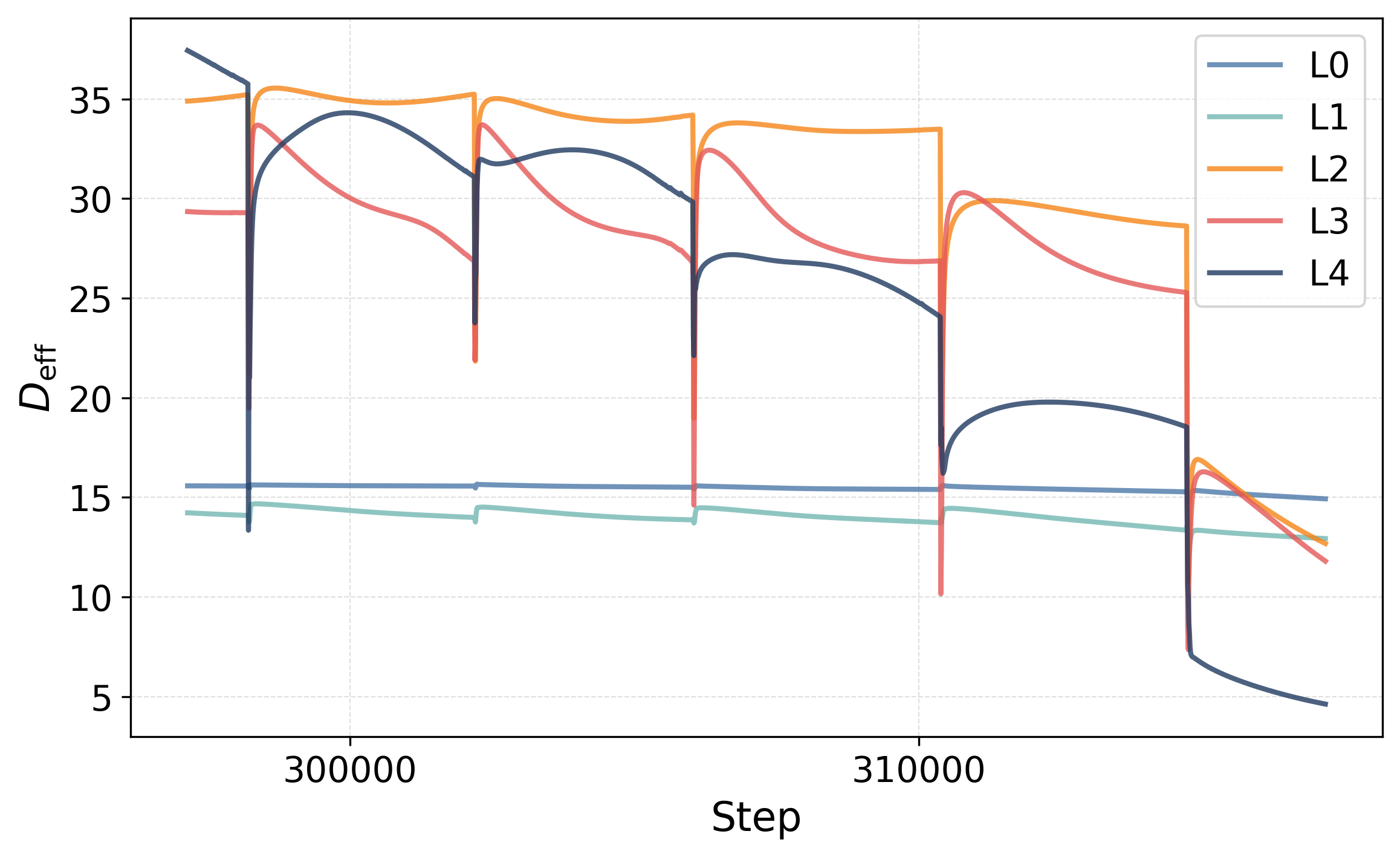}
\hfill
\includegraphics[width=0.32\textwidth]{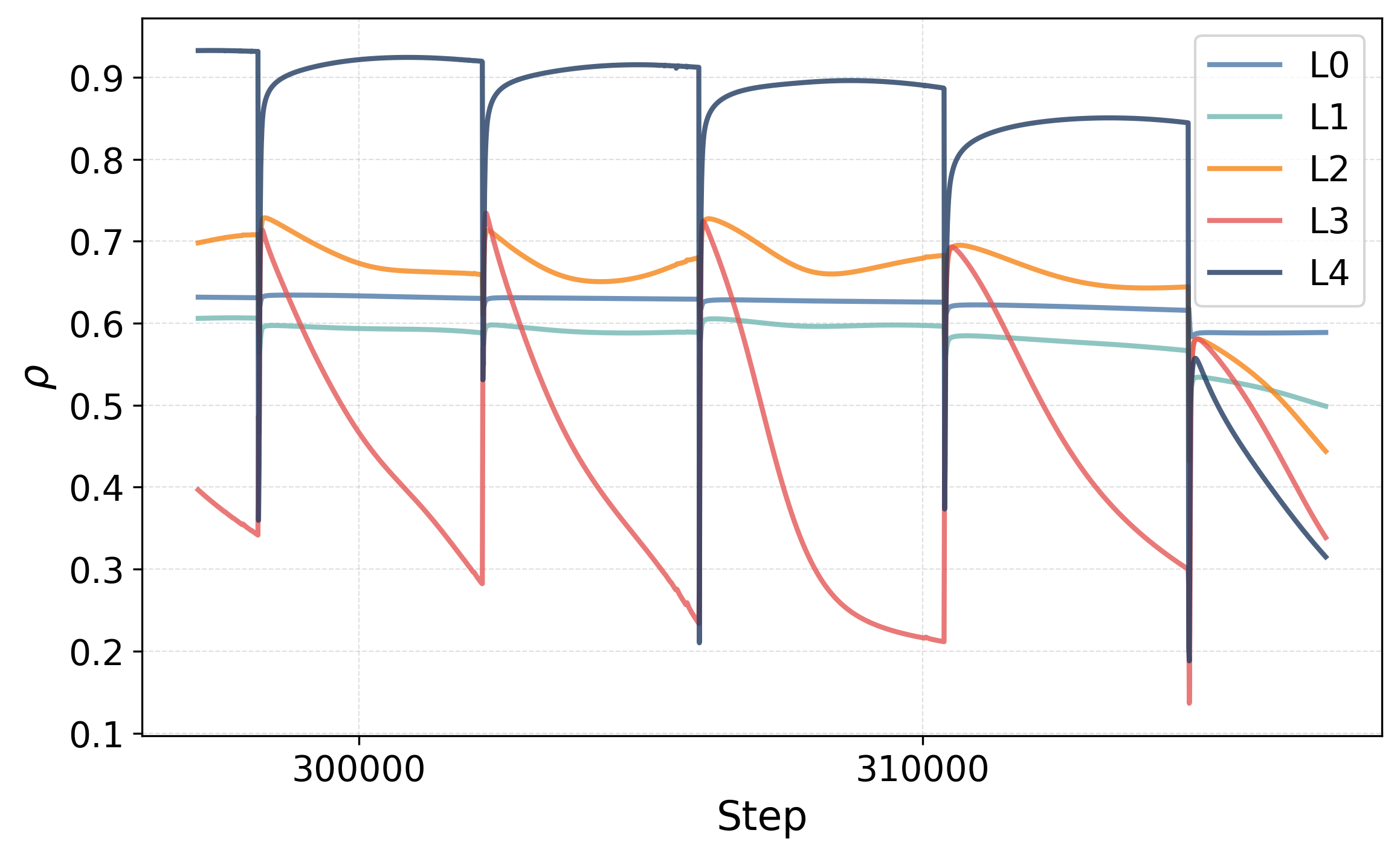}
\hfill
\includegraphics[width=0.32\textwidth]{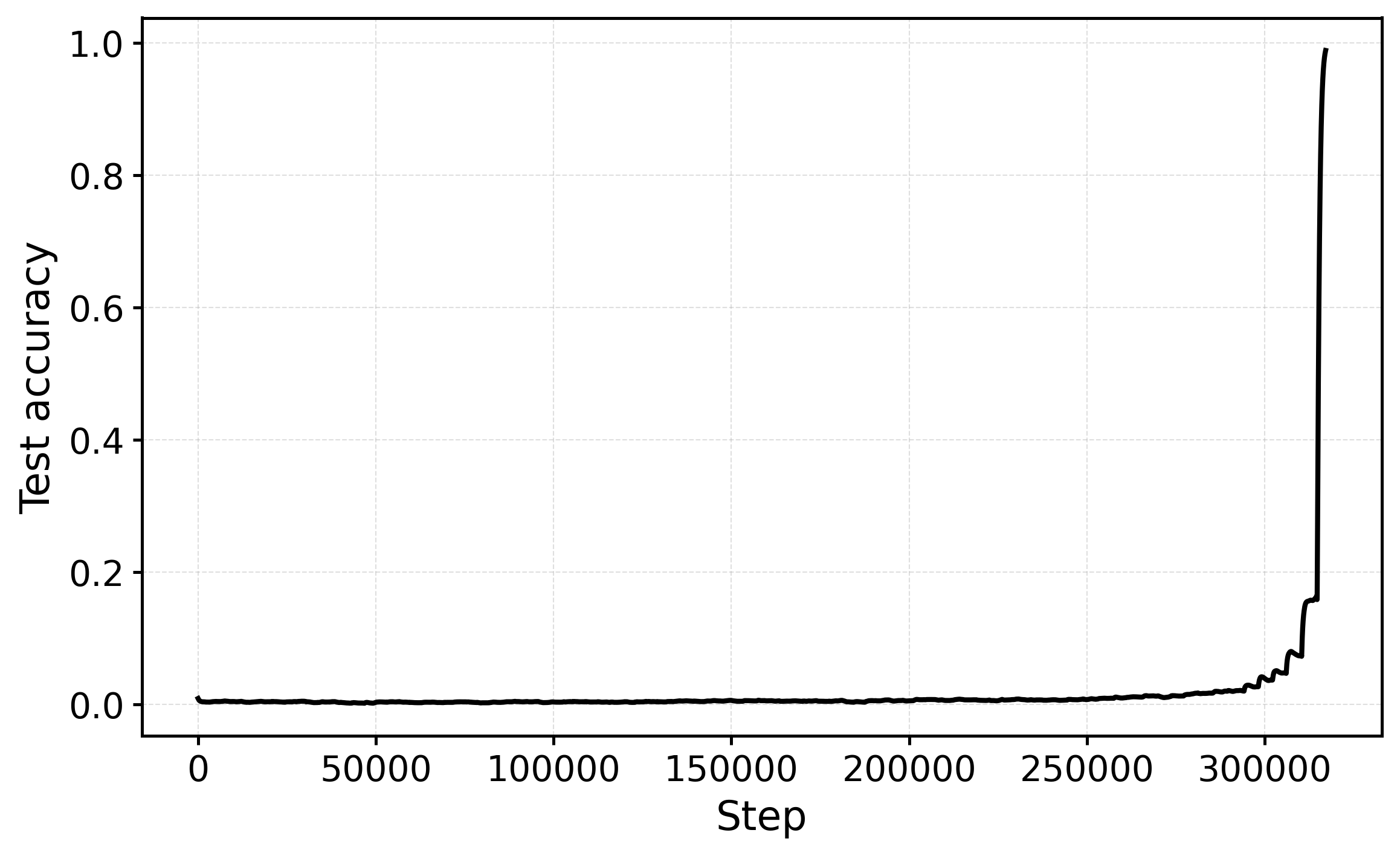}

\vspace{1mm}

{\small
(d) Final 20k steps: $D_{\mathrm{eff}}$
\hfill
(e) Final 20k steps: $\rho$
\hfill
(f) Test accuracy
}

\caption{
Evolution of accuracy and representation geometry during baseline grokking on modular addition. The model is trained without geometric regularization. Panels (a), (b), (c), and (f) show EMA-smoothed trajectories  of effective dimensionality, local neighborhood distance, train accuracy, and test accuracy. Panels (d) and (e) show the  raw geometric trajectories for dimensionality, local neighborhood distance  during the final 20K optimization steps. Delayed generalization coincides with rapid geometric reorganization, particularly in deeper layers. The deepest hidden layer ($L_4$) exhibits the largest changes in both effective dimensionality and local neighborhood distance near the grokking transition.
}
\label{fig:geometry_baseline}
\end{figure*}
\subsection{Observation}

\begin{figure*}[t]
\centering

\includegraphics[width=0.32\textwidth]{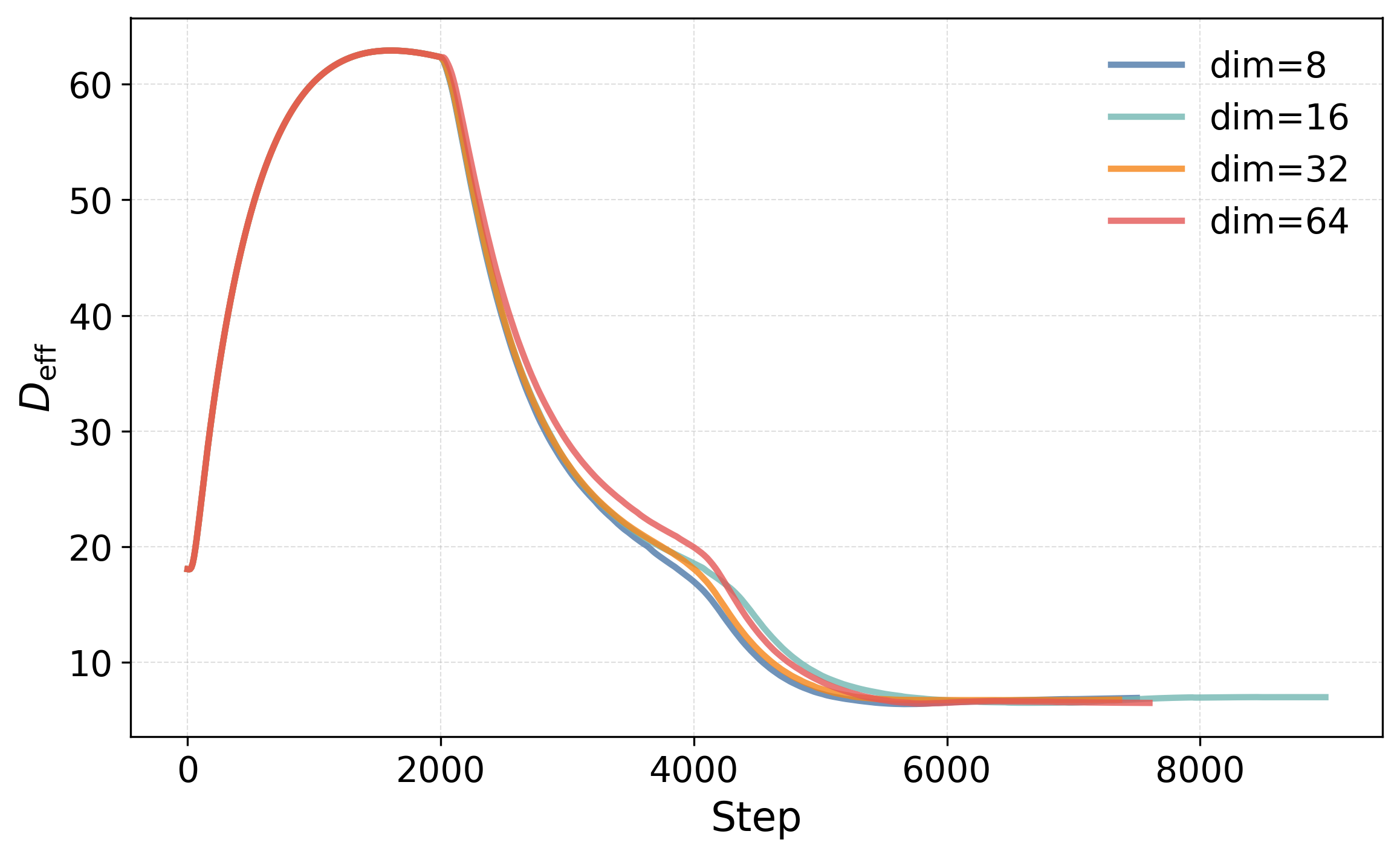}
\hfill
\includegraphics[width=0.32\textwidth]{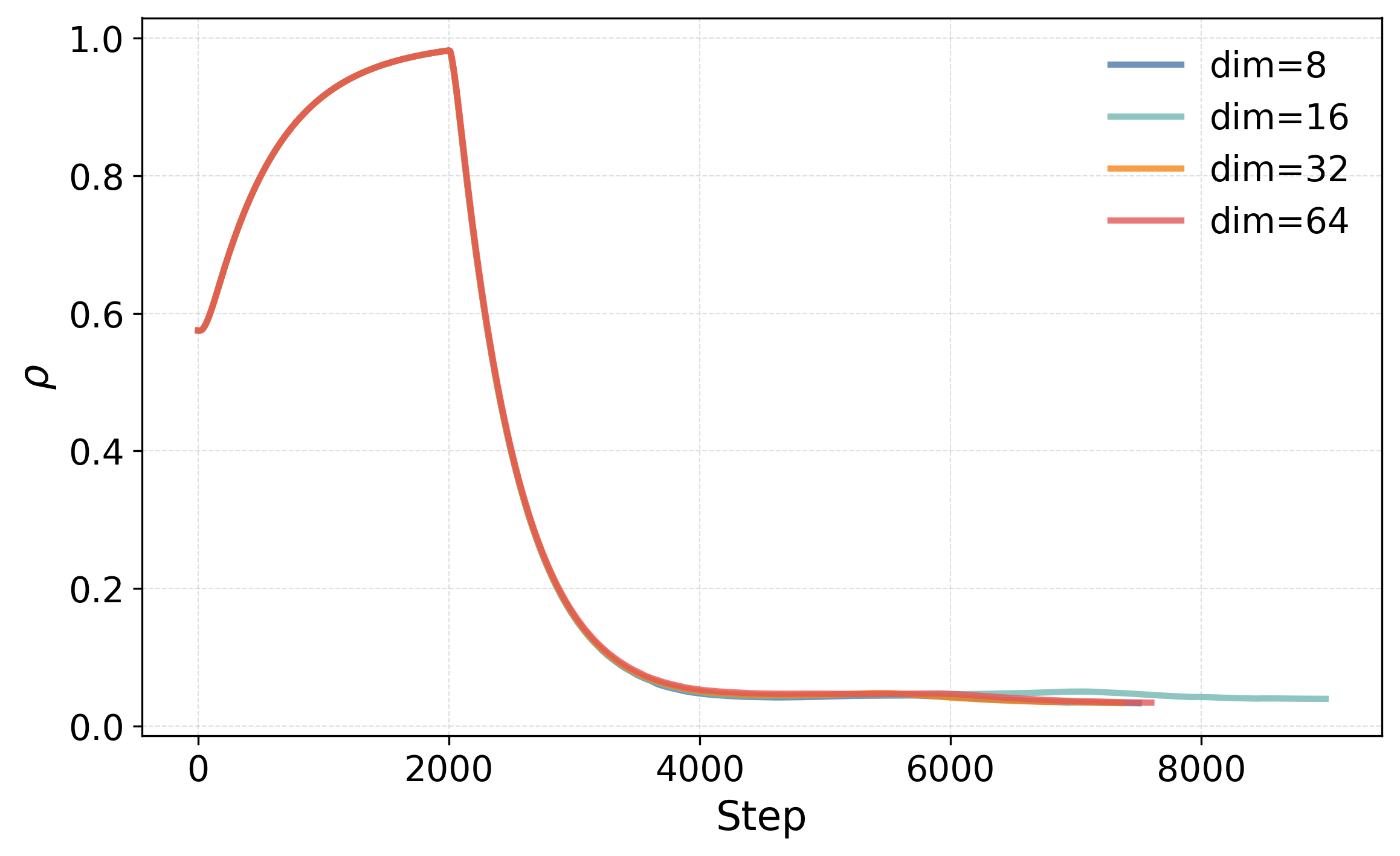}
\hfill
\includegraphics[width=0.32\textwidth]{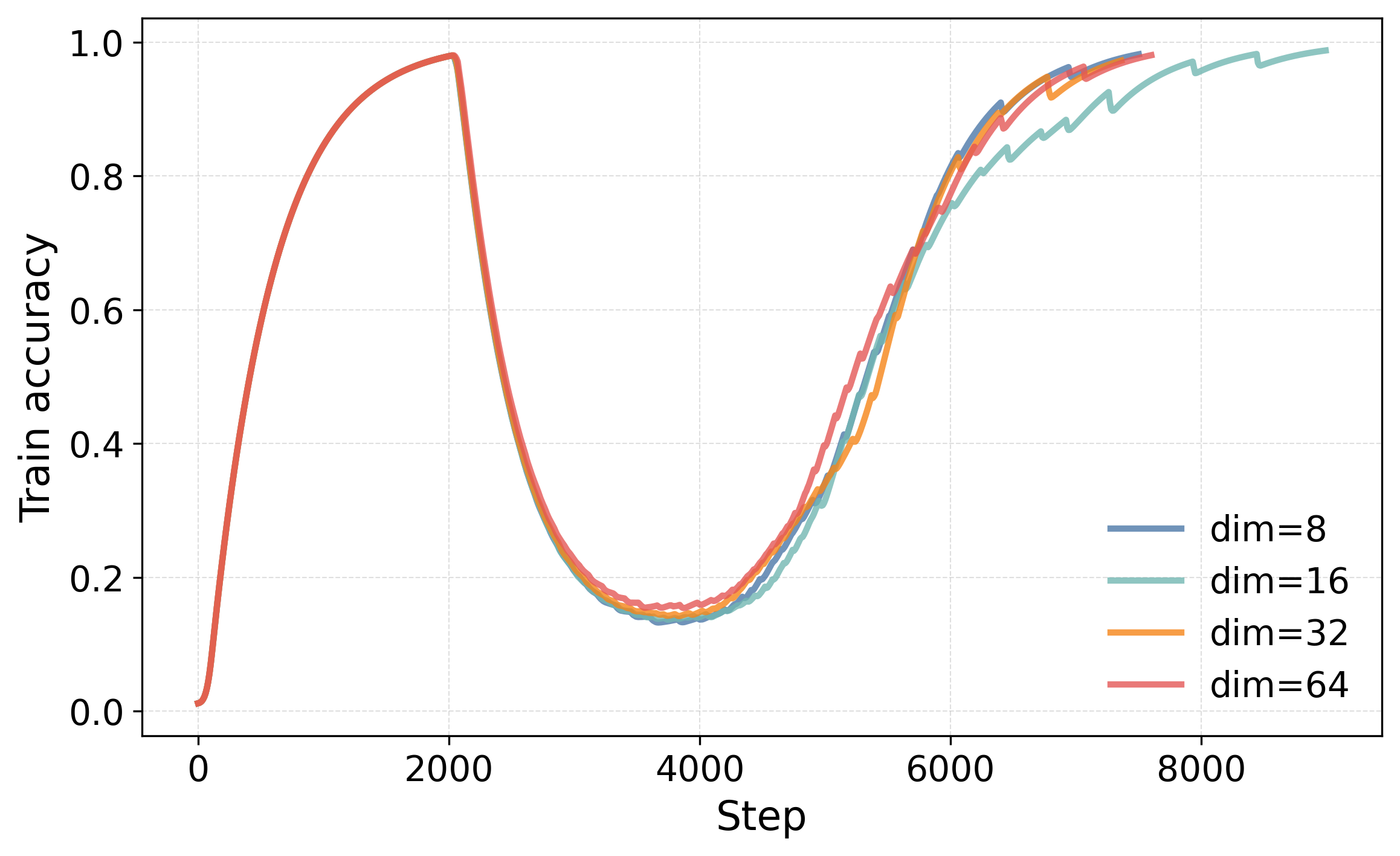}

\vspace{1mm}

{\small
(a) $L_4$: $D_{\mathrm{eff}}$
\hfill
(b) $L_4$: $\rho$
\hfill
(c) Train accuracy
}

\vspace{2mm}

\includegraphics[width=0.32\textwidth]{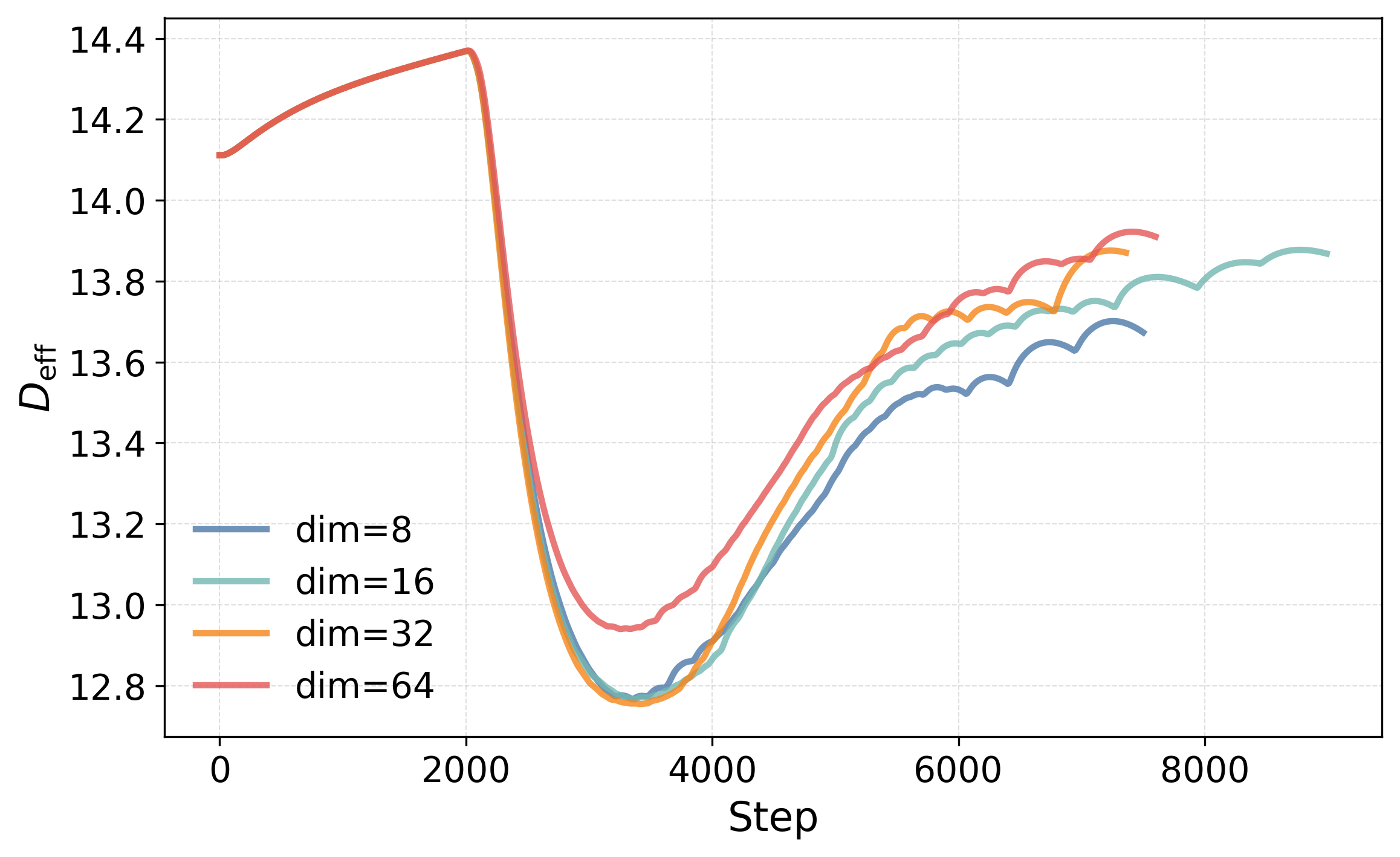}
\hfill
\includegraphics[width=0.32\textwidth]{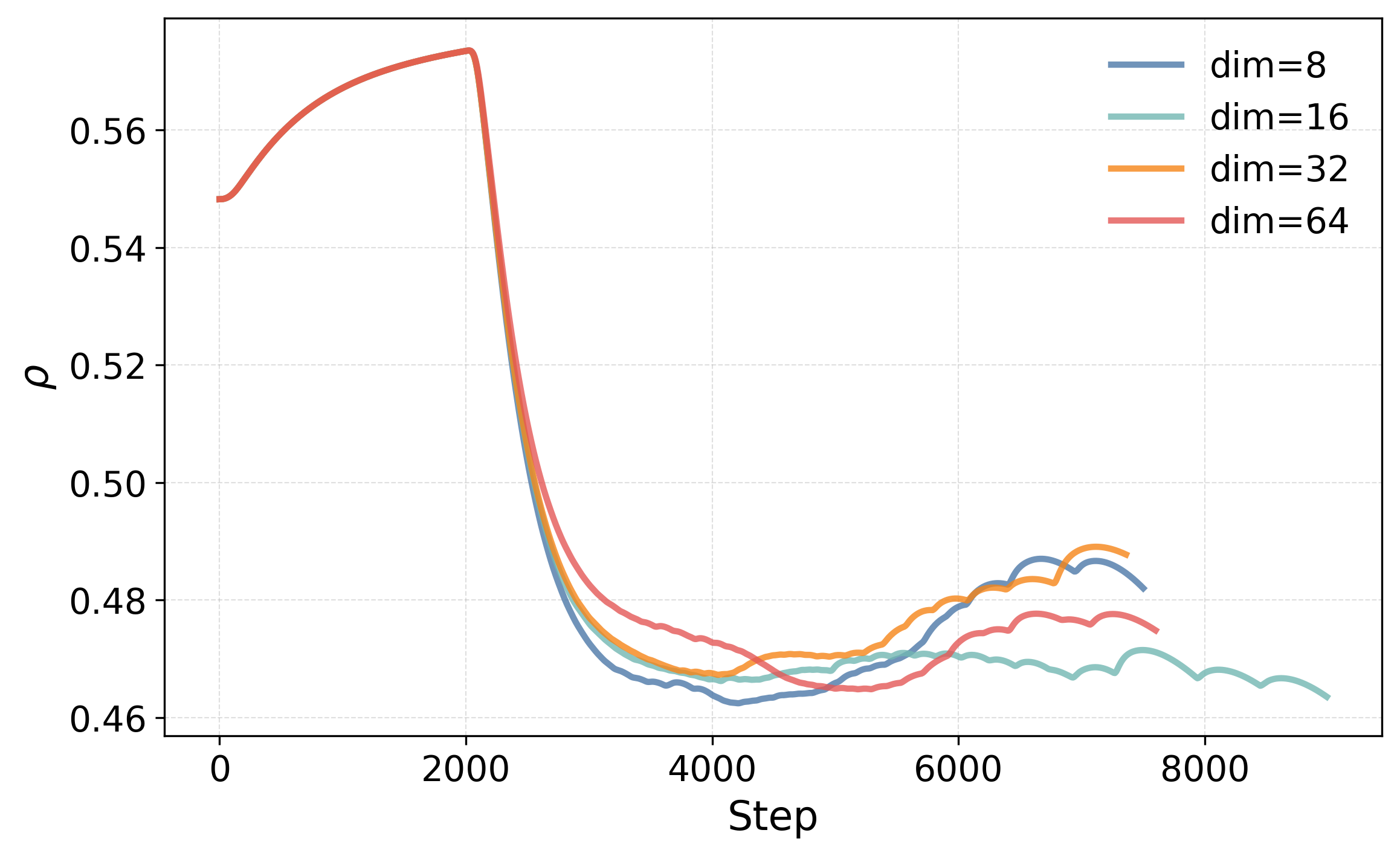}
\hfill
\includegraphics[width=0.32\textwidth]{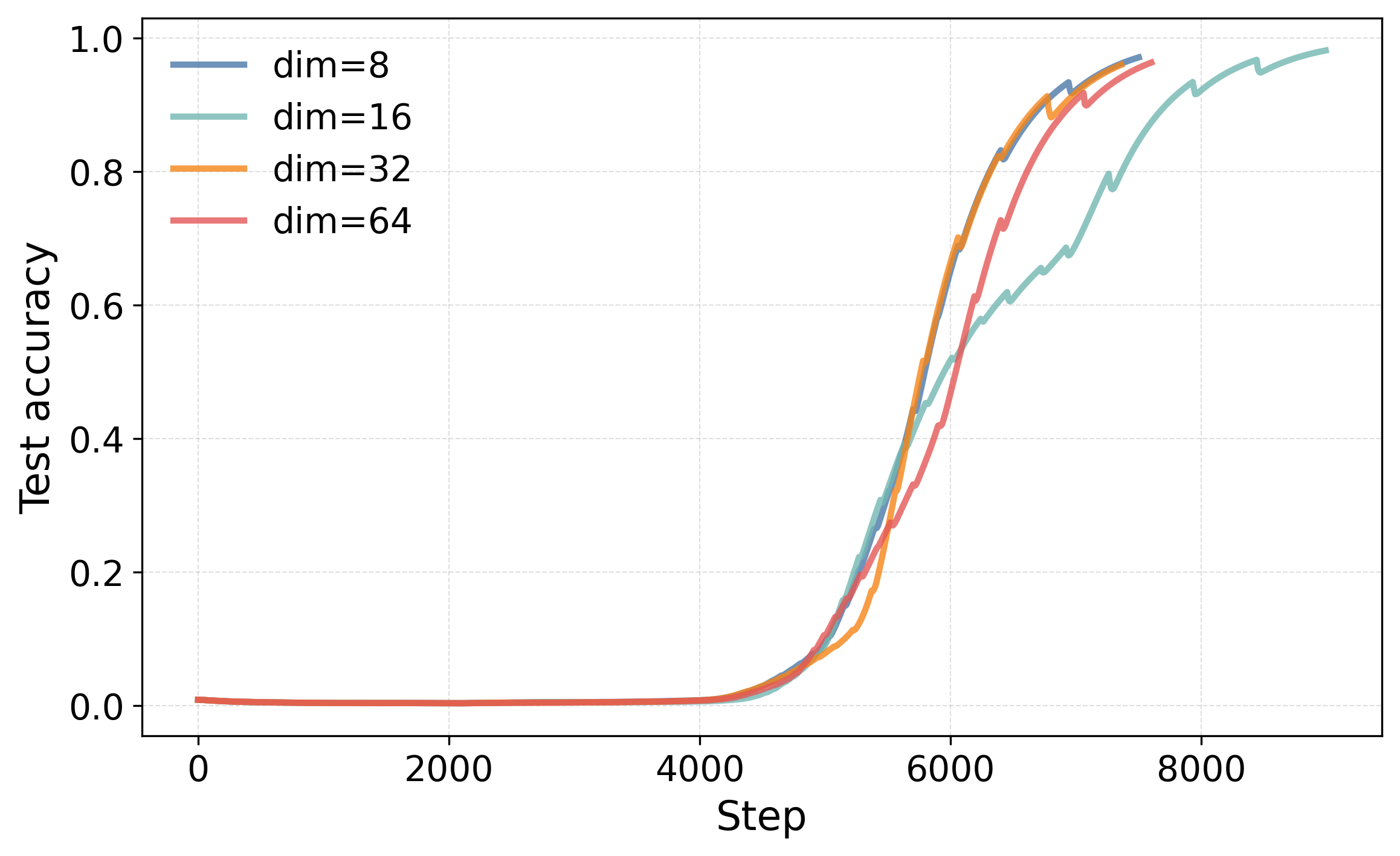}

\vspace{1mm}

{\small
(d) $L_0$: $D_{\mathrm{eff}}$
\hfill
(e) $L_0$: $\rho$
\hfill
(f) Test accuracy
}

\caption{
Evolution of accuracy and representation geometry under GeomDR for modular addition. The model is trained with different target dimensionalities $d^*$. Panels (a), (b), (c), and (f) show EMA-smoothed trajectories of effective dimensionality, local neighborhood distance, train accuracy, and test accuracy. Panels (d) and (e) show the corresponding geometric trajectories in the shallowest hidden layer. The strongest effects are observed in the deepest hidden layer ($L_4$), where dimensionality collapse occurs earliest and is followed by rapid increases in test accuracy.
}
\label{fig:dim_sweep_geometry}
\end{figure*}
We first examine the evolution of representation geometry during baseline grokking on modular addition. This baseline setting provides a reference for subsequent intervention experiments. To characterize representation geometry, we monitor effective dimensionality $D_{\mathrm{eff}}$ and the local neighborhood distance $\rho$, which measures the average distance between neighboring representations. For visualization, trajectories are smoothed with an exponential moving average (EMA) using a window of 100 steps, while the final 20,000 optimization steps are additionally shown without smoothing. 

Figure~\ref{fig:geometry_baseline} shows the evolution of effective dimensionality and local neighborhood distance across all layers. Shallow layers remain relatively stable throughout training, whereas deeper layers exhibit pronounced geometric restructuring. Effective dimensionality stays high for most of training before undergoing a sharp collapse near the grokking transition, with the strongest effect observed in the deepest hidden layer. Local neighborhood distance displays a similar pattern, remaining stable for extended periods before reorganizing rapidly near the onset of generalization. These coordinated changes occur within a narrow training interval across multiple layers, suggesting that grokking is accompanied by a rapid reorganization of representation geometry.

Across all runs analyzed and tasks considered in this study, dimensionality collapse consistently preceded grokking. In GeomDR-induced models, collapse occurred substantially earlier, reaching 50\% of its initial effective dimensionality after approximately $2.3$K optimization steps, and was followed shortly thereafter by grokking (approximately $6.8$K steps). These results indicate that interventions that accelerate grokking also accelerate dimensionality collapse, supporting the hypothesis that geometric compression is closely linked to the emergence of generalization.

The unsmoothed trajectories reveal that the transition is not continuous. Instead, grokking is associated with a small number of abrupt geometric reorganizations. These events occur simultaneously in both effective dimensionality and the local neighborhood distance $\rho$, suggesting that delayed generalization is accompanied by a large-scale restructuring of internal representations rather than gradual optimization alone. This observation motivates the hypothesis that representation geometry may play a mechanistic role in grokking dynamics. If so, then directly controlling representation geometry may alter the timing of the memorization-to-generalization transition, motivating the intervention studies presented in the following sections.

\subsection{Intervention}

To investigate the role of representation dimensionality, we introduce a regularizer that enforces a target dimensionality $d^*$. The intervention is activated at training step $t_s=2000$ and cosine-ramped to $\lambda=1$ over 333 optimization steps. We evaluate four target dimensionalities, $d^*\in\{8,16,32,64\}$, while keeping all other hyperparameters fixed. Figure~\ref{fig:dim_sweep_geometry} shows the evolution of effective dimensionality and local neighborhood distance in the shallowest ($L_0$) and deepest ($L_4$) hidden layers, together with train and test accuracy. Activation of the regularizer produces an immediate geometric transition across all measured quantities.

The strongest effect is observed in the local neighborhood distance $\rho$. Prior to the intervention, all runs follow nearly identical trajectories. Once regularization is activated, $\rho$ rapidly decreases in both shallow and deep layers, indicating substantial reorganization of the representation space. The trajectories subsequently separate according to the target dimensionality, demonstrating direct control over the resulting geometric state.

Effective dimensionality exhibits a similar response. Immediately after the intervention, $D_{\mathrm{eff}}$ decreases sharply, consistent with the suppression of variance outside the target subspace. This is followed by a recovery phase, after which larger target dimensionalities maintain higher effective dimensionality. The effect is most pronounced in the deepest layer, where the separation between settings remains visible throughout training.

Despite these substantial geometric changes, train and test accuracy remain broadly similar across settings. Together, these results demonstrate that dimensionality regularization provides a direct mechanism for controlling representation geometry during training.

\subsection{Control: Regularization Schedule and Target Dimensionality}

\begin{figure}[t]
\centering
\includegraphics[width=\columnwidth]{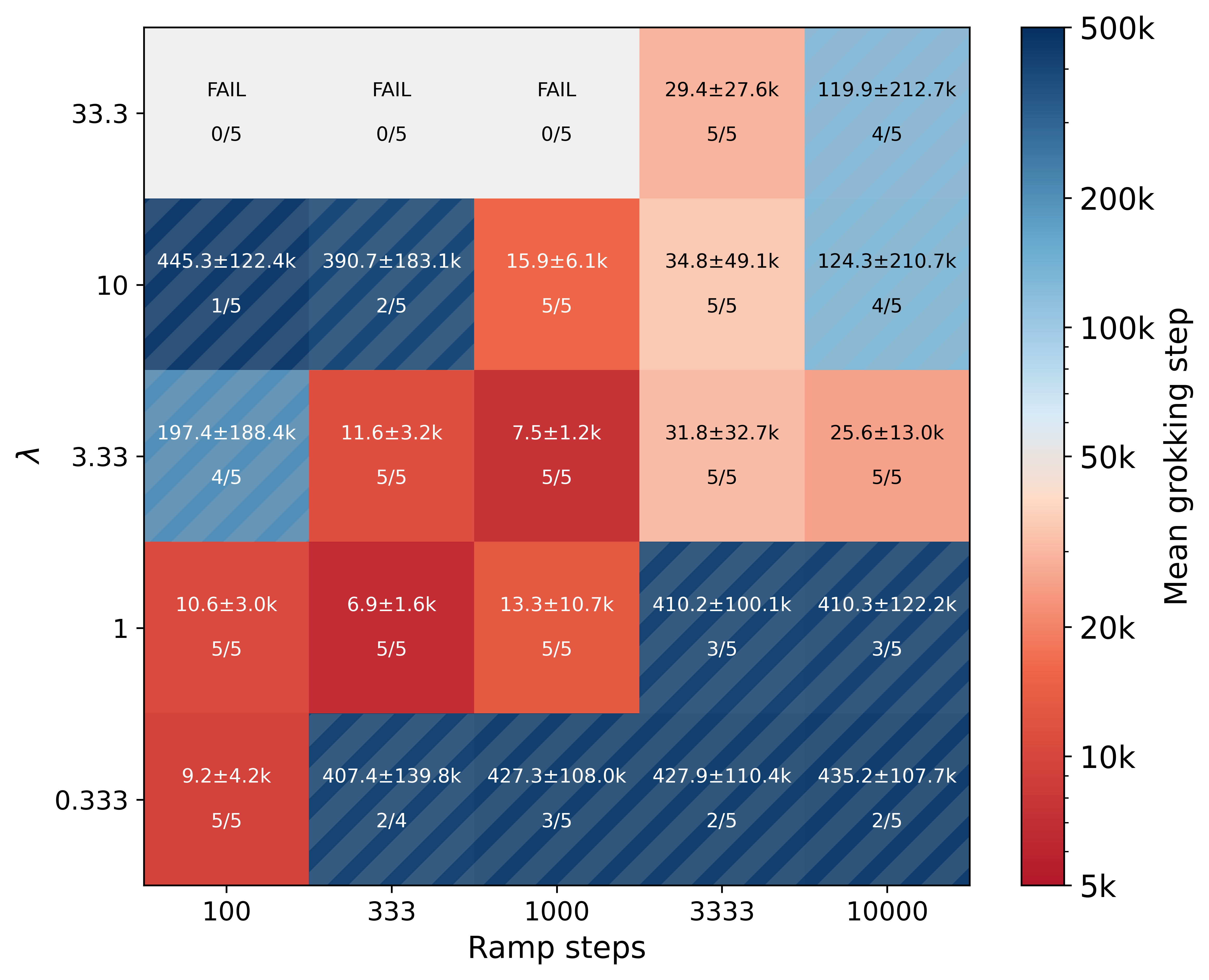}
\caption{
Schedule sweep for modular addition ($d^*=16$). Cells report the mean grokking step (K) $\pm$ one standard deviation; fractions indicate successful runs. Grey cells denote failed configurations (at least one failed run). Intermediate $\lambda_{\max}$ and $T_{\mathrm{ramp}}$ produce the fastest grokking, with $\lambda_{\max}=1$ and $T_{\mathrm{ramp}}=333$ performing best.
}
\label{fig:schedule_sweep}
\end{figure}

To identify an effective intervention schedule, we fixed the target dimensionality to $d^*=16$ and performed a grid search over the final regularization strength $\lambda_{\max}$ and ramp duration $T_{\mathrm{ramp}}$. Figure~\ref{fig:schedule_sweep} summarizes the results. Relative to the baseline ( $362.1\pm111.0$K steps), GeomDR substantially accelerates grokking across a broad range of schedules, with the best configuration reaching successful generalization after approximately 7K steps. Intermediate regularization strengths and ramp durations perform best, whereas very weak interventions have little effect and overly strong interventions can destabilize training or prevent grokking. Based on these results, we select $\lambda_{\max}=1$ and $T_{\mathrm{ramp}}=333$ for subsequent experiments.

\begin{figure}[t]
\centering
\includegraphics[width=\columnwidth]{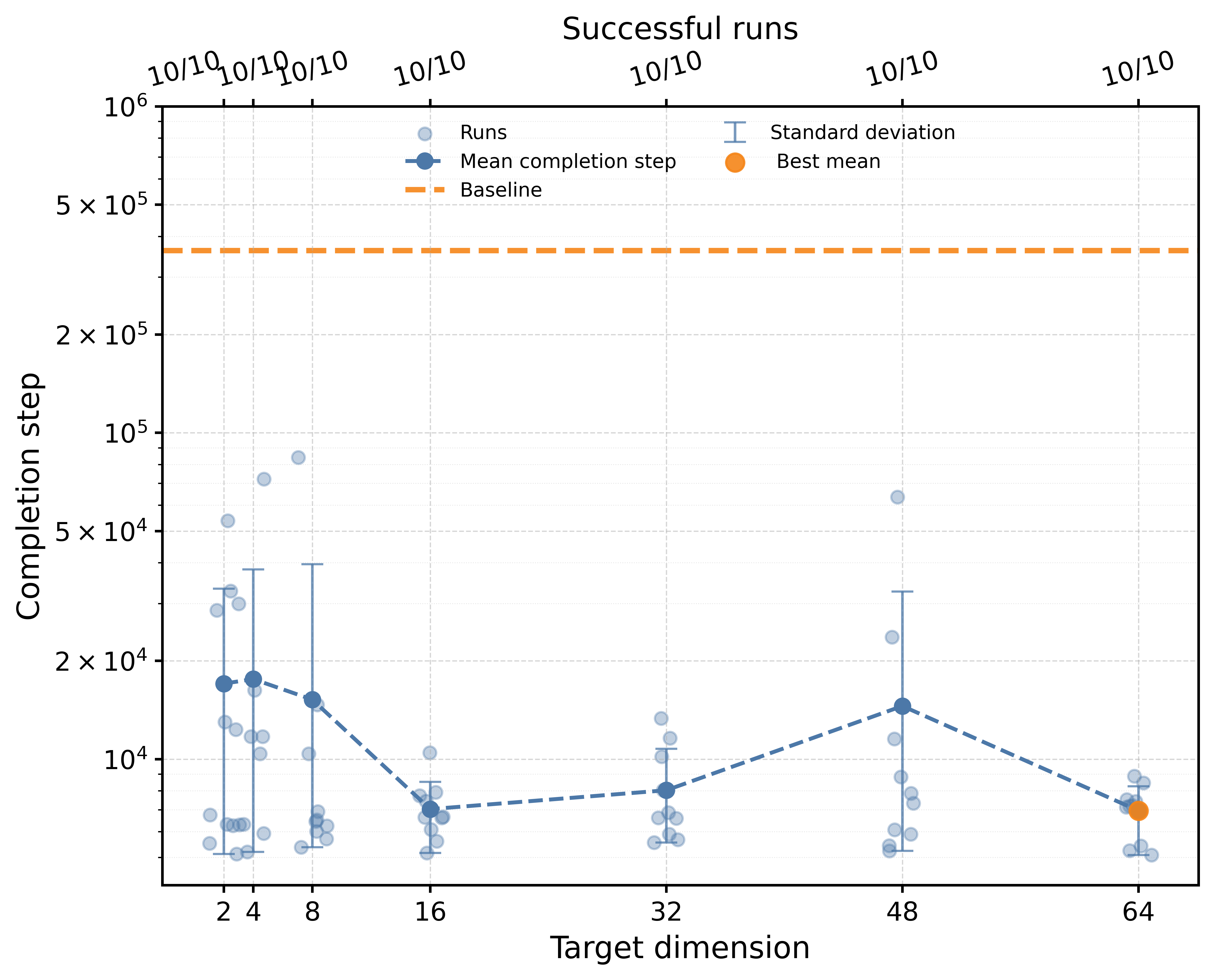}
\caption{
Effect of target dimensionality $d^*$ on grokking. Points show individual runs, the dashed curve the mean, and error bars one standard deviation. The orange dashed line denotes the baseline. Moderate and large target dimensions ($16\leq d^*\leq64$) substantially accelerate grokking, whereas aggressive compression ($d^*\leq8$) slows and destabilizes learning.
}
\label{fig:dimensionality_sweep}
\end{figure}

Using this schedule, we vary the target dimensionality over $d^*\in\{2,4,8,16,32,48,64\}$ and evaluate each configuration across ten random seeds. As shown in Figure~\ref{fig:dimensionality_sweep}, aggressive compression ($d^*\leq8$) produces slower and more variable grokking, whereas dimensions in the range $16\leq d^*\leq64$ consistently yield strong acceleration. The best mean performance is obtained at $d^*=64$, reaching the grokking criterion after approximately 7K steps. Similar trends are observed for modular division and permutation composition (Appendix~\ref{app:mlp_ablations}).

\subsection{Ablation Studies: Intervention Timing}

\begin{figure}[t]
\centering
\includegraphics[width=\columnwidth]{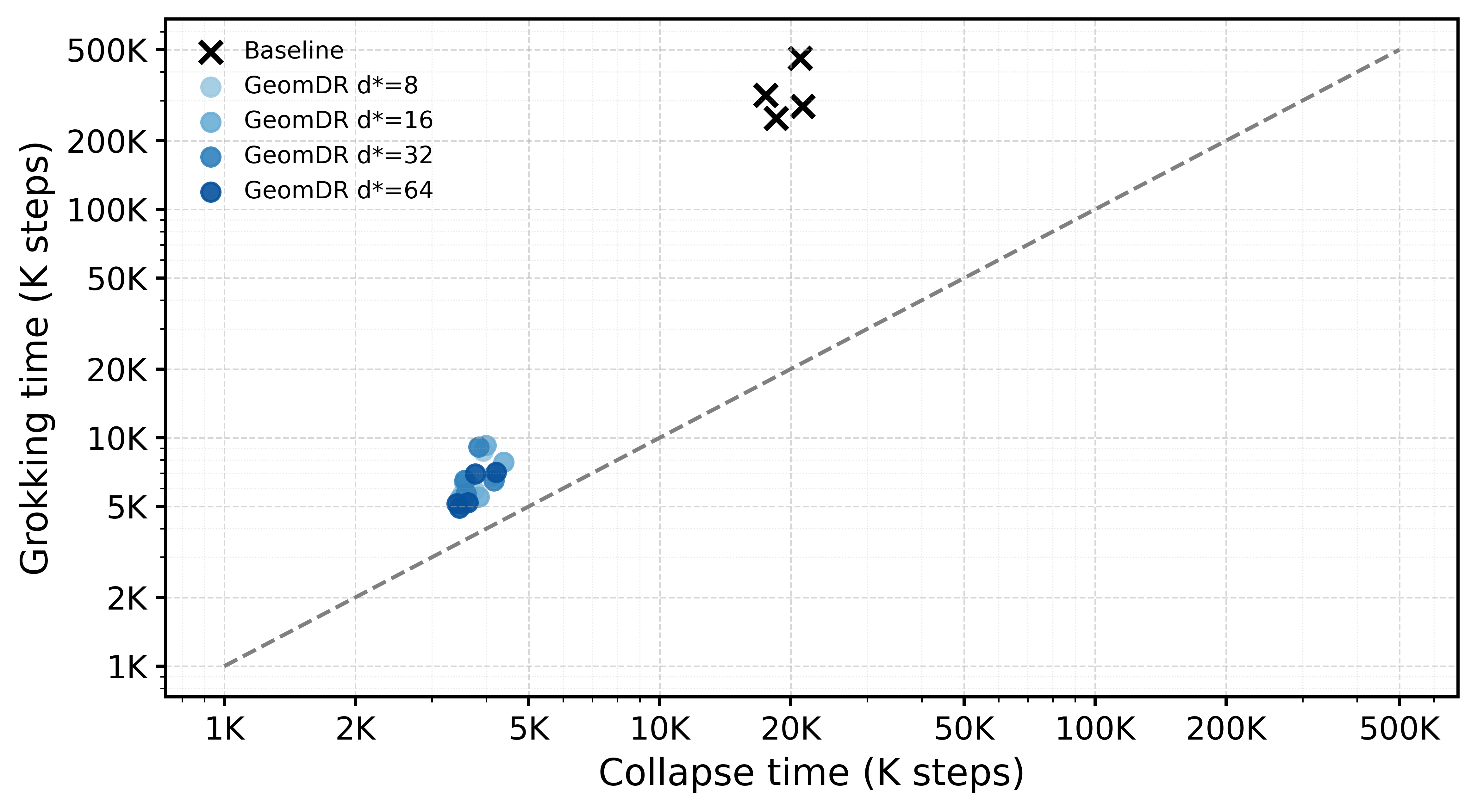}

\caption{
Relationship between representation collapse time and grokking time. Each point corresponds to one run. The dashed diagonal denotes $t_{\mathrm{collapse}}=t_{\mathrm{grok}}$. 
}
\label{fig:collapse_vs_grokking}
\end{figure}

To investigate the relationship between dimensionality collapse and grokking, we measured the collapse time of the deepest hidden representation as the first training step at which its effective dimensionality fell below 50\% of its initial value. We then compared this quantity to the grokking time, defined as the first step at which the grokking criterion above is satisfied. Figure~\ref{fig:collapse_vs_grokking} shows that dimensionality collapse consistently precedes grokking in both baseline and GeomDR models. In the baseline setting, collapse occurs relatively early in training (approximately 15--25K steps), whereas grokking emerges much later (approximately 250--500K steps), producing a large temporal gap between geometric compression and successful generalization. GeomDR shifts both events to substantially earlier stages of training, with collapse occurring after only 3--5K steps and grokking following shortly thereafter at approximately 5--10K steps. Consequently, the intervention not only accelerates dimensionality collapse but also markedly reduces the lag between collapse and generalization. Across all target dimensionalities and random seeds, GeomDR moves the system closer to the diagonal $t_{\mathrm{collapse}}=t_{\mathrm{grok}}$, indicating a substantially tighter coupling between geometric reorganization and the onset of generalization.

\begin{figure}[t]
\centering
\includegraphics[width=\linewidth]{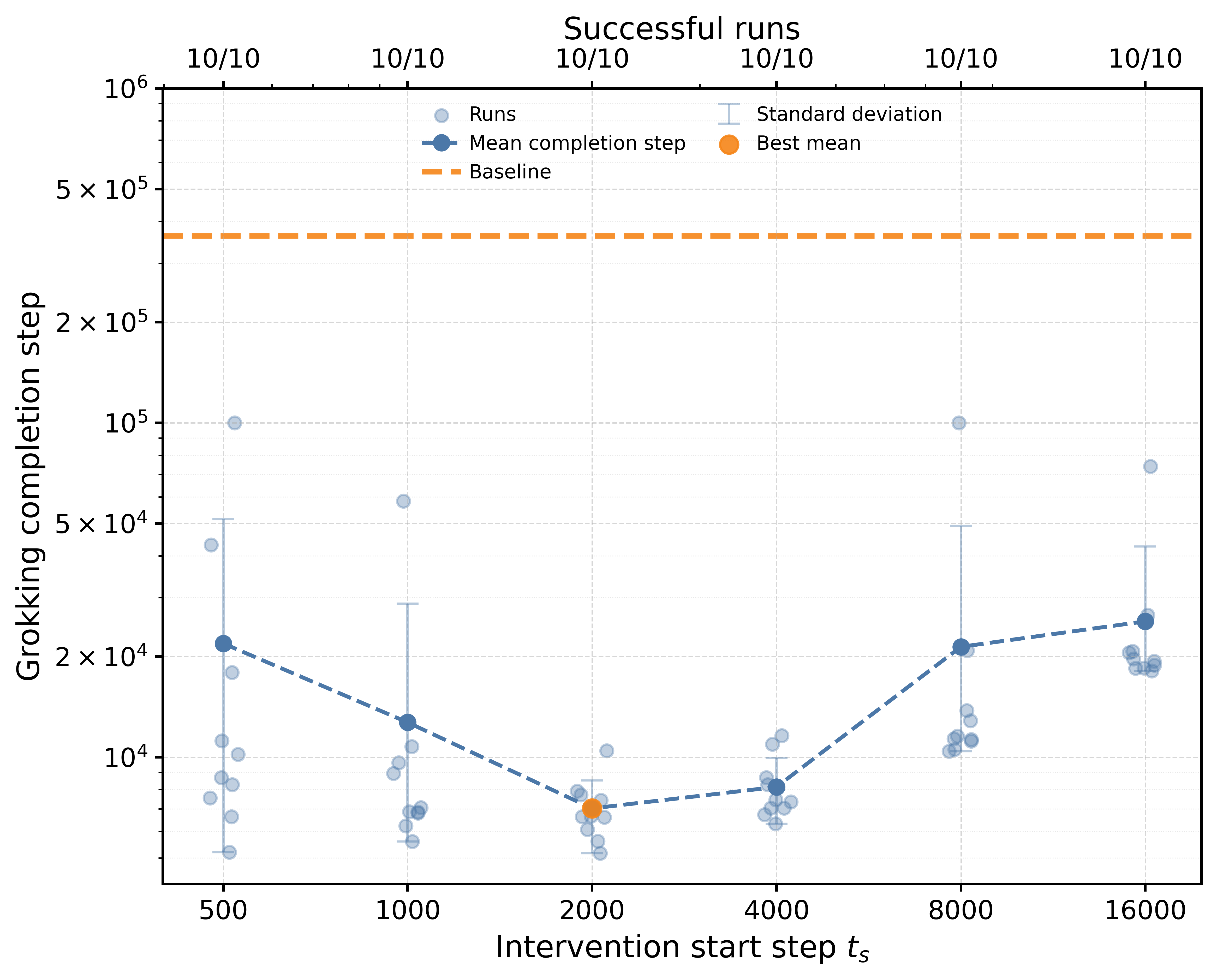}
\caption{
Intervention-start sweep for modular addition. Points show individual runs; the dashed curve and shaded region denote the mean and one standard deviation. Intermediate activation times yield the fastest grokking.
}

\label{fig:switch_sweep}
\end{figure}
The effectiveness of geometric regularization depends strongly on intervention timing. Figure~\ref{fig:switch_sweep} shows the relationship between the intervention-start step $t_s$ and the resulting grokking step. A clear U-shaped trend is observed. Very early interventions ($t_s=500$ and $t_s=1000$) exhibit substantially slower and more variable convergence across random seeds. The fastest and most stable grokking behavior occurs for intermediate activation times ($t_s=2000$--$4000$), where the mean grokking step is minimized. Delaying the intervention beyond this region progressively increases the number of optimization steps required for successful generalization.

These results suggest the existence of a critical temporal window during which geometric regularization is most effective. Applying the intervention too early appears to disrupt the formation of useful task representations, whereas applying it too late reduces its ability to influence the memorization-to-generalization transition. Overall, the results support the hypothesis that geometric interventions are most beneficial after an initial unconstrained learning phase but before memorizing solutions become firmly established.

Additional intervention-start sweeps and implementation details are provided in Appendix~\ref{app:mlp_ablations}. Intervention timing substantially influences performance, although the optimal activation step varies across tasks and architectures.

\begin{table*}[t]
\centering
\caption{
Summary of the best GeomDR configuration for each task and architecture.
Values denote mean grokking step $\pm$ standard deviation (K optimization steps).
S/T indicates runs satisfying the grokking criterion.
Unsuccessful runs were assigned the maximum training budget (500K for MLPs, 200K for Transformers) when computing statistics.
}
\label{tab:main_summary}

\footnotesize
\setlength{\tabcolsep}{2.5pt}

\begin{tabular}{lccc ccc ccc}
\toprule

&
\multicolumn{3}{c}{Addition}
&
\multicolumn{3}{c}{Division}
&
\multicolumn{3}{c}{Permutation}
\\

\cmidrule(lr){2-4}
\cmidrule(lr){5-7}
\cmidrule(lr){8-10}

Architecture
&
Step
&
S/T
&
Imp.
&
Step
&
S/T
&
Imp.
&
Step
&
S/T
&
Imp.
\\

\midrule

MLP
&
362.1 $\pm$ 111.0
&
4/5
&
--
&
330.9 $\pm$ 100.9
&
5/5
&
--
&
300.2 $\pm$ 207.0
&
3/5
&
--
\\

MLP + GeomDR
&
\textbf{6.9 $\pm$ 1.6}
&
5/5
&
\textbf{52.5$\times$}
&
\textbf{8.4 $\pm$ 5.0}
&
5/5
&
\textbf{39.4$\times$}
&
\textbf{12.8 $\pm$ 1.7}
&
5/5
&
\textbf{23.5$\times$}
\\

\midrule

Transformer
&
66.4 $\pm$ 12.0
&
5/5
&
--
&
89.4 $\pm$ 27.3
&
5/5
&
--
&
72.6 $\pm$ 33.0
&
5/5
&
--
\\

Transformer + GeomDR
&
\textbf{33.0 $\pm$ 2.5}
&
5/5
&
\textbf{2.01$\times$}
&
\textbf{48.7 $\pm$ 8.8}
&
5/5
&
\textbf{1.84$\times$}
&
\textbf{48.9 $\pm$ 9.1}
&
5/5
&
\textbf{1.48$\times$}
\\

\bottomrule
\end{tabular}
\end{table*}

\subsection{Generalization Across Tasks and Architectures}

We report grokking times, success rates, and relative improvements for all evaluated settings. The relationship between representation geometry and grokking remains consistent across all evaluated settings. Table~\ref{tab:main_summary} summarizes the best-performing GeomDR configuration for each task and architecture.

For MLPs, GeomDR produces large and consistent improvements across all tasks, reducing the mean grokking time from $362.1$K to $6.9$K steps on modular addition ($52.5\times$), from $330.9$K to $8.4$K steps on modular division ($39.4\times$), and from $300.2$K to $12.8$K steps on permutation learning ($23.5\times$). In all cases, GeomDR achieves a 100\% success rate. For Transformers, the effect remains consistent but more moderate. GeomDR reduces grokking time from $66.4$K to $33.0$K steps on modular addition ($2.01\times$), from $89.4$K to $48.7$K steps on modular division ($1.84\times$), and from $72.6$K to $48.9$K steps on permutation learning ($1.48\times$), while maintaining a 100\% success rate across all tasks. Additional ablation studies, hyperparameter sweeps, and architecture-specific analyses are provided in Appendix~\ref{app:mlp_ablations}. Together, these results indicate that GeomDR accelerates grokking across diverse tasks and architectures, although the magnitude of improvement varies across model classes.

\section{Discussion}

Our results suggest that representation geometry can serve as an effective control signal for grokking. Across a range of tasks, schedules, dimensionalities, and architectures, GeomDR consistently influences delayed generalization. In many MLP settings, the intervention accelerates grokking by more than an order of magnitude, while Transformer experiments show smaller but qualitatively similar improvements. The smaller gains observed in Transformers may reflect their substantially faster baseline grokking dynamics, leaving less room for acceleration than in MLPs.

The geometric analyses reveal systematic changes in effective dimensionality and local distance near the grokking transition. Across all successful runs, dimensionality collapse consistently preceded grokking, and interventions that accelerated grokking also accelerated the onset of collapse. These observations suggest that delayed generalization is accompanied by a substantial reorganization of internal representations. Previous work on grokking has largely treated representation geometry and compression as emergent consequences of learning \cite{power2022grokking,liu2022towards,varma2023grokking,nanda2023progress}. In contrast, our results show that directly modifying representation geometry is sufficient to substantially alter delayed generalization dynamics. While this does not establish representation dimensionality as the sole causal mechanism underlying grokking, it demonstrates that geometric interventions are capable of systematically altering when grokking occurs. Geometric interventions provide a new experimental methodology for studying learning dynamics and suggest that representation geometry may serve as a controllable degree of freedom for studying optimization and generalization.  Rather than passively observing representational changes during training, researchers can directly manipulate geometric properties and measure the resulting effects on optimization and generalization. Importantly, the observed acceleration is not restricted to a single task or finely tuned hyperparameter configuration. Effective interventions are obtained across a broad range of target dimensionalities and activation schedules, indicating that the relationship between representation geometry and grokking is robust rather than task-specific. 
Several limitations remain. First, our experiments focus on small-scale algorithmic grokking benchmarks, and it remains unclear whether similar geometric interventions are effective in larger-scale domains such as language or vision. Second, while GeomDR provides a direct geometric intervention, the precise mechanisms linking representation geometry and delayed generalization remain an open theoretical question.  Appendix~\ref{appendix:spectral}
provides a geometric interpretation of the regularizer in terms
of effective-rank control, low rank approximation, and
representation volume compression.

\section{Conclusion}

Across all tasks and architectures studied, dimensionality collapse consistently preceded generalization. Motivated by this observation, we introduced Geometric Dimensionality Regularization (GeomDR), a spectral regularizer that directly controls the effective dimensionality of hidden representations. Across diverse settings, GeomDR substantially alters grokking dynamics, accelerating, delaying, or suppressing generalization. These results suggest that representation geometry is not merely a correlate of grokking but a useful target for intervention.

\clearpage
\bibliography{bibliography}

\clearpage
\onecolumn

\renewcommand{\contentsname}{Contents}

{
\hypersetup{linkcolor=black}

\Large
\setlength{\cftbeforesecskip}{8pt}
\setlength{\cftbeforesubsecskip}{3pt}
\tableofcontents
}

\hypersetup{linkcolor=red}

\clearpage
\twocolumn

\appendix

\section*{Appendix}
\addcontentsline{toc}{section}{Appendix}

\setcounter{secnumdepth}{1}

\section{Spectral Properties of GeomDR}
\label{appendix:spectral}
The results in this appendix characterize the optimization objective induced by Geometric Dimensionality Regularization (GeomDR) and establish several geometric properties directly encouraged by the regularizer, including effective-rank control, low rank covariance approximation, and representation volume compression. Importantly, these results do not constitute a formal proof of why grokking accelerates under GeomDR. Rather, they describe the geometric structures that GeomDR explicitly promotes during optimization. Our working hypothesis is that accelerated grokking emerges because GeomDR reduces the space of high-dimensional memorizing solutions and biases optimization toward more compact representations, a mechanism that is consistent with the empirical observations presented in the main paper.

Let \(Z \in \mathbb{R}^{N \times d}\) denote a matrix of hidden representations and let

\begin{equation}
C
=
\frac{1}{N-1}
(Z-\bar Z)^\top
(Z-\bar Z)
\label{eq:covariance}
\end{equation}

denote the corresponding covariance matrix. Let

\begin{equation}
\mu_1
\ge
\mu_2
\ge
\cdots
\ge
\mu_d
\ge
0
\label{eq:eigenvalues}
\end{equation}

be the eigenvalues of \(C\).

The Geometric Dimensionality Regularization (GeomDR) objective is

\begin{equation}
L_{\mathrm{GeomDR}}
=
\sum_{i=d^*+1}^{d}
\mu_i
\label{eq:geomdr}
\end{equation}

where \(d^*\) is the target dimensionality.

Using

\begin{equation}
\mathrm{Tr}(C)
=
\sum_{i=1}^{d}
\mu_i
\label{eq:trace}
\end{equation}

the objective can be rewritten as
\begin{equation}
L_{\mathrm{GeomDR}}
=
\mathrm{Tr}(C)
-
\sum_{i=1}^{d^*}
\mu_i
\label{eq:geomdr_trace}
\end{equation}

This formulation admits several useful interpretations.

\paragraph{Proposition 1 (Effective Rank Control).}

For \(\epsilon \in (0,1)\), define the effective rank

\begin{equation}
r_\epsilon(C)
=
\min
\left\{
k :
\frac{
\sum_{i=1}^{k}\mu_i
}{
\sum_{i=1}^{d}\mu_i
}
\ge
1-\epsilon
\right\}
\label{eq:effective_rank}
\end{equation}

If

\begin{equation}
L_{\mathrm{GeomDR}}
\le
\epsilon\,\mathrm{Tr}(C)
\label{eq:effective_rank_condition}
\end{equation}

then

\begin{equation}
r_\epsilon(C)
\le
d^*
\label{eq:effective_rank_bound}
\end{equation}

\paragraph{Proof.}

Since \eqref{eq:geomdr_trace} holds, the assumption
\eqref{eq:effective_rank_condition} implies

\begin{equation}
\sum_{i=1}^{d^*}\mu_i
\ge
(1-\epsilon)\,\mathrm{Tr}(C)
\label{eq:variance_capture_bound}
\end{equation}

Hence the leading \(d^*\) eigenvalues explain at least a fraction \(1-\epsilon\) of the total variance. By definition of \(r_\epsilon(C)\),

\begin{equation}
r_\epsilon(C)
\le
d^*.
\end{equation}

\hfill $\square$

This proposition formalizes the fact that GeomDR directly limits the number of statistically significant covariance directions.
\paragraph{Proposition 2 (Low Rank Approximation Bound).}

Let \(C_{d^*}\) denote the optimal rank-\(d^*\) approximation of \(C\), and let
\(\|\cdot\|_F\) denote the Frobenius norm. Then

\begin{equation}
\|C-C_{d^*}\|_F
\le
L_{\mathrm{GeomDR}}
\label{eq:low_rank_bound}
\end{equation}
\paragraph{Proof.}

By the Eckart--Young theorem \cite{eckart1936approximation},

\begin{equation}
\|C-C_{d^*}\|_F^2
=
\sum_{i=d^*+1}^{d}
\mu_i^2
\label{eq:frobenius_error}
\end{equation}

Since all eigenvalues are nonnegative,

\begin{equation}
\sum_{i=d^*+1}^{d}
\mu_i^2
\le
\left(
\sum_{i=d^*+1}^{d}
\mu_i
\right)^2
=
L_{\mathrm{GeomDR}}^2.
\end{equation}

Taking square roots yields \eqref{eq:low_rank_bound}.

\hfill $\square$

Thus minimizing GeomDR minimizes an upper bound on the reconstruction error of the best rank-\(d^*\) covariance approximation.

\paragraph{Proposition 3 (Representation Volume Compression).}

Assume that the nonzero covariance spectrum of a representation matrix
is

\begin{equation}
\mu_1 \ge \mu_2 \ge \cdots \ge \mu_r > 0,
\end{equation}

where \(r=\mathrm{rank}(C)\). The covariance ellipsoid associated
with \(C\) has volume proportional to the square root of the determinant
of \(C\) \cite{jolliffe2002principal},

\begin{equation}
\mathrm{Vol}(C)
\propto
\sqrt{\det(C)}
=
\prod_{i=1}^{r}\sqrt{\mu_i}.
\end{equation}

\begin{equation}
L_{\mathrm{GeomDR}}
=
\sum_{i=d^*+1}^{d}\mu_i
\le \epsilon,
\end{equation}

then every tail eigenvalue satisfies

\begin{equation}
\mu_i \le \epsilon,
\qquad
i>d^*.
\end{equation}

Consequently,

\begin{equation}
\begin{aligned}
\mathrm{Vol}(C)
&=
\left(
\prod_{i=1}^{d^*}\sqrt{\mu_i}
\right)
\left(
\prod_{i=d^*+1}^{r}\sqrt{\mu_i}
\right)
\\
&\le
\left(
\prod_{i=1}^{d^*}\sqrt{\mu_i}
\right)
\left(
\prod_{i=d^*+1}^{r}\sqrt{\epsilon}
\right)
\\
&\le
\left(
\prod_{i=1}^{d^*}\sqrt{\mu_i}
\right)
\epsilon^{(r-d^*)/2}.
\end{aligned}
\end{equation}

Thus, as \(L_{\mathrm{GeomDR}} \rightarrow 0\), the representation
volume outside the leading \(d^*\)-dimensional subspace vanishes.

\paragraph{Proof.}

Since

\begin{equation}
\sum_{i=d^*+1}^{d}\mu_i
\le \epsilon
\end{equation}

and all eigenvalues are nonnegative,

\begin{equation}
\mu_i \le \epsilon,
\qquad
i>d^*.
\end{equation}

Therefore,

\begin{equation}
\prod_{i=d^*+1}^{r}\sqrt{\mu_i}
\le
\prod_{i=d^*+1}^{r}\sqrt{\epsilon}
=
\epsilon^{(r-d^*)/2}.
\end{equation}

Substituting into the volume expression yields

\begin{equation}
\mathrm{Vol}(C)
\le
\left(
\prod_{i=1}^{d^*}\sqrt{\mu_i}
\right)
\epsilon^{(r-d^*)/2}.
\end{equation}

\hfill $\square$

The empirical results suggest that GeomDR accelerates grokking by reducing the geometric complexity of learned representations. Neural networks can often fit the training data using a large family of distinct solutions, many of which distribute information across numerous weakly informative directions in representation space. Such solutions achieve low training error but need not capture the underlying algorithmic structure of the task.

In contrast, successful generalization on algorithmic tasks may require the discovery of more structured representations. For modular arithmetic and related symbolic problems, these representations often appear substantially more concentrated, with variance captured by a relatively small number of dominant latent directions. Under this view, memorizing and algorithmic solutions occupy regions of representation space with different geometric characteristics: memorizing solutions tend to be spectrally diffuse and high-dimensional, whereas algorithmic solutions appear more spectrally concentrated.

GeomDR explicitly suppresses covariance mass outside a target subspace through the objective \eqref{eq:geomdr}.

Propositions~1--3 show that minimizing this objective simultaneously controls the effective rank of the representation covariance, improves its low rank approximation, and compresses the geometric volume associated with tail covariance directions. In particular, Proposition~3 implies that as \(L_{\mathrm{GeomDR}}\) decreases, the volume contributed by dimensions outside the leading \(d^*\)-dimensional subspace vanishes.

Consequently, optimization is no longer free to spread information across a large number of weak directions. Instead, representations are encouraged to concentrate variance into a smaller set of dominant components, reducing the geometric volume available to spectrally diffuse solutions. This may substantially restrict the family of memorizing representations accessible during training and bias optimization toward more compact solutions.

Under this hypothesis, GeomDR does not directly create algorithmic representations. Rather, it modifies the geometry of the optimization landscape by making high-dimensional, spectrally diffuse solutions increasingly expensive. As training proceeds, optimization is therefore encouraged to move toward compressed representations that may be more likely to capture task structure. This shift can reduce the time required to transition from memorization to generalization, leading to earlier grokking.

The broad dimensionality plateaus observed in our experiments are consistent with this interpretation. Acceleration occurs across a wide range of target dimensionalities rather than only at a single finely tuned value, suggesting that the critical factor is the suppression of excess representational degrees of freedom rather than the precise dimensionality itself.

While this explanation remains a hypothesis rather than a formal proof, it provides a geometric interpretation that is consistent with the observed reductions in effective dimensionality, increased spectral concentration, covariance-volume compression, and substantially earlier grokking transitions across multiple tasks and architectures.

\clearpage
\section{Tasks and Architectures}
\label{app:tasks_architectures}
\subsection{Modular Addition}

The modular addition task receives a pair of integers
$(a,b)\in\mathbb{Z}_{97}^{2}$ and predicts

\[
y=(a+b)\bmod 97.
\]

The dataset contains all possible ordered pairs in
$\mathbb{Z}_{97}^{2}$, resulting in
$97^{2}=9409$
examples.

\subsection{Modular Division}

The modular division task receives a pair
$(a,b)$ with $b\neq0$ and predicts

\[
y=(a\cdot b^{-1})\bmod 97,
\]

where $b^{-1}$ denotes the multiplicative inverse of
$b$ in $\mathbb{Z}_{97}$.

The dataset contains all valid pairs
$(a,b)$ with
$a\in\mathbb{Z}_{97}$
and
$b\in\{1,\ldots,96\}$,
resulting in $97 \times 96 = 9312$ examples.

\subsection{Permutation Composition}

To evaluate whether the observed effects extend beyond modular arithmetic, we additionally consider permutation composition. Let $S_5$ denote the symmetric group on five elements.
Given two permutations
$\sigma,\tau\in S_5$,
the task is to predict their composition

\[
y=\sigma\circ\tau,
\]

defined by

\[
(\sigma\circ\tau)(i)
=
\sigma\!\bigl(\tau(i)\bigr).
\]

Since $|S_5|=120$, the dataset contains $120^{2}=14400$ ordered pairs of permutations.

\subsection{MLP Architecture}

Each input token is mapped to an
$8$-dimensional embedding vector.
The embeddings corresponding to the two inputs are concatenated,

\[
z_0=[e(a);e(b)],
\]

yielding a $16$-dimensional input representation.

The concatenated representation is projected into a hidden space of dimension $128$:

\[
h_0
=
\mathrm{LayerNorm}
\!\left(
\mathrm{GELU}(W_{\mathrm{in}}z_0+b_{\mathrm{in}})
\right).
\]

The network then applies three residual feed-forward blocks

\[
h_{l+1}
=
\mathrm{LayerNorm}
\!\left(
h_l+
\mathrm{GELU}(W_l h_l+b_l)
\right),
\]
\[
\qquad l\in\{0,1,2\}.
\]

The final hidden representation is mapped to the output space using a linear classifier

\[
\hat y
=
W_{\mathrm{out}}h_3+b_{\mathrm{out}}.
\]

The architecture parameters are summarized below:

\begin{itemize}
\item Embedding dimension: $8$
\item Input dimension: $16$
\item Hidden dimension: $128$
\item Residual blocks: $3$
\item Activation: GELU
\item Normalization: LayerNorm
\item Output dimension: $97$ (arithmetic tasks), $120$ (permutation composition)
\end{itemize}

Geometric Dimensionality Regularization (GeomDR) is applied to the hidden representations produced by the MLP. Specifically, the regularization term is evaluated for the outputs of all hidden layers $(h_0,h_1,h_2,h_3)$ and summed across layers. The input embedding representation $z_0$ is excluded from the regularization.

\paragraph{Permutation Composition MLP.}

For the permutation composition task, we use a larger MLP configuration.
The task involves predicting one of $|S_5|=120$ possible permutation classes,
compared with 97 output classes in the modular arithmetic tasks. The
embedding dimension is increased to $16$ and the hidden dimension is increased
to $256$ while preserving the same residual MLP structure.

The permutation MLP architecture uses:

\begin{itemize}
\item Embedding dimension: $16$
\item Input dimension: $32$
\item Hidden dimension: $256$
\item Residual blocks: $3$
\item Activation: GELU
\item Normalization: LayerNorm
\item Output dimension: $120$
\end{itemize}

For permutation composition experiments, we use
$d^*=48$ to account for the larger hidden representation dimension.

All MLP models are trained using the same optimization protocol. We use full-batch AdamW optimization with a learning rate of $10^{-3}$ and weight decay of $10^{-1}$.

\subsection{Transformer Architecture}

To evaluate whether the proposed geometric intervention generalizes beyond multilayer perceptrons, we additionally consider a Transformer encoder architecture \cite{vaswani2017attention}.

Each input symbol is mapped to a learned embedding of dimension

\[
d_{\mathrm{model}} = 32.
\]

All tasks considered in this work operate on ordered pairs of input symbols. Consequently, each example is represented as a sequence of length two. Learned positional embeddings are added to the token embeddings before being processed by the Transformer encoder.

The encoder consists of two Transformer layers. Each layer employs multi-head self-attention with

\[
n_{\mathrm{heads}} = 4
\]

attention heads and a feed-forward network of dimension

\[
d_{\mathrm{ff}} = 64.
\]

The architecture uses GELU activations, residual connections, LayerNorm, and the pre-normalization formulation of the Transformer. Dropout is disabled in all experiments.

Let

\[
H^{(0)}
=
E(X)+P
\]

denote the input token embeddings together with the learned positional embeddings. The Transformer encoder computes

\[
H^{(\ell+1)}
=
\mathrm{TransformerLayer}
\!\left(
H^{(\ell)}
\right),
\]

for

\[
\ell \in \{0,1\},
\]

where \(L=2\) denotes the number of encoder layers.

After the final encoder layer, token representations are aggregated using mean pooling across the sequence dimension,

\[
h
=
\frac{1}{2}
\sum_{i=1}^{2}
H_i^{(L)}.
\]

The pooled representation is subsequently normalized using LayerNorm and mapped to the output space through a linear classifier,

\[
\hat{y}
=
W_{\mathrm{out}} h
+
b_{\mathrm{out}}.
\]

The architecture parameters are summarized below:

\begin{itemize}
\item Embedding dimension ($d_{\mathrm{model}}$): $32$
\item Attention heads: $4$
\item Transformer layers: $2$
\item Feed-forward dimension: $64$
\item Activation: GELU
\item Normalization: LayerNorm (pre-normalization)
\item Pooling: mean pooling
\item Output dimension: task dependent
\end{itemize}

Transformer models for modular addition and modular division are trained using full-batch AdamW optimization with a learning rate of $10^{-3}$ and weight decay of $10^{-1}$. For permutation composition, we use a learning rate of $3\times10^{-4}$ and weight decay of $1.0$.

For the permutation composition task, the default Transformer configuration was insufficient to reliably exhibit grokking behavior within the available training budget with and without interventions. To obtain a stable grokking regime suitable for evaluating geometric interventions, we increased the model capacity only for the permutation task. Specifically, permutation composition experiments use

\[
d_{\mathrm{model}} = 64,
\qquad
L = 3,
\qquad
d_{\mathrm{ff}} = 128.
\]

All other architectural and optimization settings remain unchanged. This larger Transformer is used for both the baseline and GeomDR experiments on permutation composition. The larger architecture was introduced solely to obtain a reproducible grokking regime and is not used for the addition or division tasks, which reliably grok under the default Transformer configuration.

Geometric Dimensionality Regularization (GeomDR) is applied to the hidden token representations produced by each Transformer layer. For each layer, token representations from all examples and sequence positions are collected into a matrix

\[
\mathbb{R}^{N_{\mathrm{tokens}} \times d_{\mathrm{model}}},
\]

from which the covariance matrix is computed.

The GeomDR penalty is then evaluated by summing the eigenvalues beyond a target dimensionality \(d^*\). The total regularization term is obtained by summing the GeomDR penalties across all Transformer layers.

For each task, the train--test split is generated once and kept fixed across all runs. For all tasks, 30\% of examples are used for training and
70\% for testing. Different random seeds therefore affect only model initialization and optimizer state. Reported results are averaged across seeds, and variability is reported as one standard deviation. This protocol follows the standard experimental setting commonly adopted in grokking studies \cite{power2022grokking,liu2022towards}.

\noindent

\begin{table*}[t]
\centering

\caption{
Ablation studies of Geometric Dimensionality Regularization (GeomDR) on three grokking tasks using the MLP architecture. The table reports dimension sweeps (varying the target dimensionality $d^*$) and intervention-time sweeps (varying the GeomDR activation step $t_s$). Values denote the mean grokking step $\pm$ one standard deviation (in thousands of optimization steps). Baseline results are computed from 5 random seeds, whereas all GeomDR sweeps are computed from 10 random seeds. The baseline corresponds to standard training without Geometric Dimensionality Regularization (GeomDR). Success/Total columns report the number of runs that satisfied the grokking criterion out of all runs. For runs that did not satisfy the criterion within the training budget, the maximum budget of 500K optimization steps was used when computing the reported mean and standard deviation.
}

\label{tab:mlp_ablations}

\begin{tabular}{lcc cc cc}
\toprule
&
\multicolumn{2}{c}{Addition}
&
\multicolumn{2}{c}{Division}
&
\multicolumn{2}{c}{Permutation}
\\

\cmidrule(lr){2-3}
\cmidrule(lr){4-5}
\cmidrule(lr){6-7}

Setting
&
Step (K)
&
Success/Total
&
Step (K)
&
Success/Total
&
Step (K)
&
Success/Total
\\

\midrule

Baseline
&
$362.1 \pm 111.0$
&
$4/5$
&
$330.9\pm 100.9$
&
$5/5$
&
$300.2 \pm 207.0$
&
$3/5$
\\

\midrule
\multicolumn{7}{c}{\textbf{Dimension Sweep}}
\\
\midrule

2
&
$17.0 \pm 16.3$
&
$10/10$
&
$81.1 \pm 154.3$
&
$9/10$
&
$209.3 \pm 216.1$
&
$7/10$
\\

4
&
$17.6 \pm 20.5$
&
$10/10$
&
$62.3 \pm 154.0$
&
$9/10$
&
$183.8 \pm 221.2$
&
$7/10$
\\

8
&
$15.2 \pm 24.3$
&
$10/10$
&
$64.8 \pm 110.8$
&
$10/10$
&
$176.6 \pm 189.6$
&
$9/10$
\\

16
&
$7.0 \pm 1.5$
&
$10/10$
&
$37.1 \pm 41.0$
&
$10/10$
&
$\mathbf{176.5 \pm 182.2}$
&
$\mathbf{9/10}$
\\

32
&
$8.0 \pm 2.7$
&
$10/10$
&
$\mathbf{19.5 \pm 15.5}$
&
$10/10$
&
$217.2 \pm 216.4$
&
$7/10$
\\

48
&
$14.5 \pm 18.1$
&
$10/10$
&
$61.9 \pm 154.1$
&
$9/10$
&
$176.5 \pm 198.4$
&
$8/10$
\\

64
&
$\mathbf{7.0 \pm 1.3}$
&
$10/10$
&
$60.9 \pm 154.5$
&
$9/10$
&
$181.3 \pm 198.3$
&
$9/10$
\\

\midrule
\multicolumn{7}{c}{\textbf{Intervention Time Sweep}}
\\
\midrule

500
&
$21.9 \pm 29.6$
&
$10/10$
&
$15.3 \pm 6.5$
&
$10/10$
&
$160.3 \pm 198.6$
&
$9/10$
\\

1000
&
$12.7 \pm 16.1$
&
$10/10$
&
$\mathbf{13.3 \pm 7.3}$
&
$10/10$
&
$149.3 \pm 196.5$
&
$8/10$
\\

2000
&
$\mathbf{7.0 \pm 1.5}$
&
$10/10$
&
$61.9 \pm 154.1$
&
$9/10$
&
$176.5 \pm 198.4$
&
$8/10$
\\

4000
&
$8.1 \pm 1.8$
&
$10/10$
&
$66.4 \pm 153.1$
&
$9/10$
&
$163.3 \pm 171.6$
&
$9/10$
\\

8000
&
$21.4 \pm 27.8$
&
$10/10$
&
$62.8 \pm 153.6$
&
$9/10$
&
$\mathbf{162.1 \pm 155.1}$
&
$\mathbf{10/10}$
\\

16000
&
$25.5 \pm 17.2$
&
$10/10$
&
$24.5 \pm 8.8$
&
$10/10$
&
$273.6 \pm 231.8$
&
$7/10$
\\

\bottomrule
\end{tabular}
\end{table*}
\FloatBarrier
\section{Ablations}
\label{app:ablations}
\subsection{MLP}
\label{app:mlp_ablations}

\begin{figure*}[!t]
\centering


\includegraphics[width=0.40\textwidth]{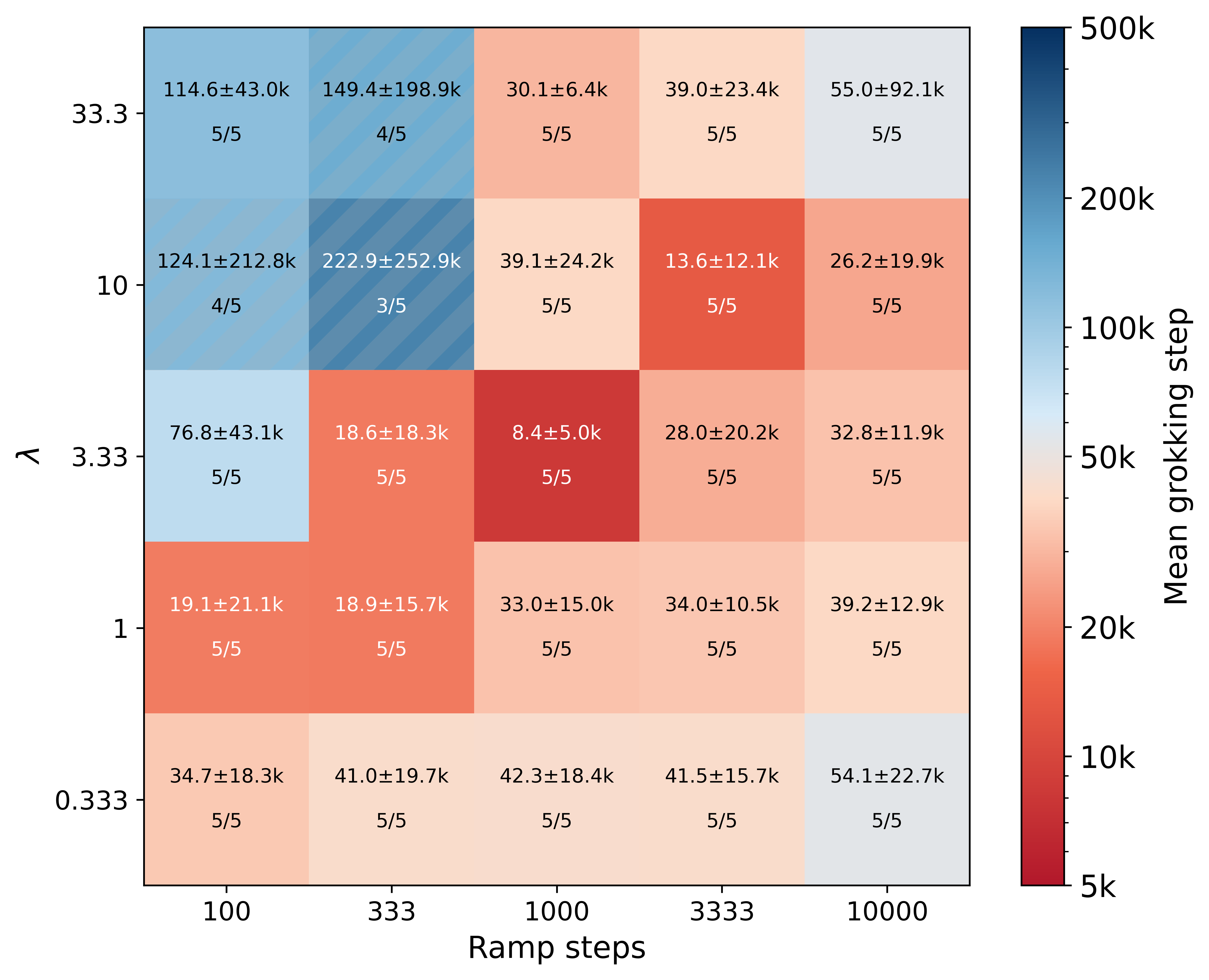}
\hspace{0.005\textwidth}
\includegraphics[width=0.40\textwidth]{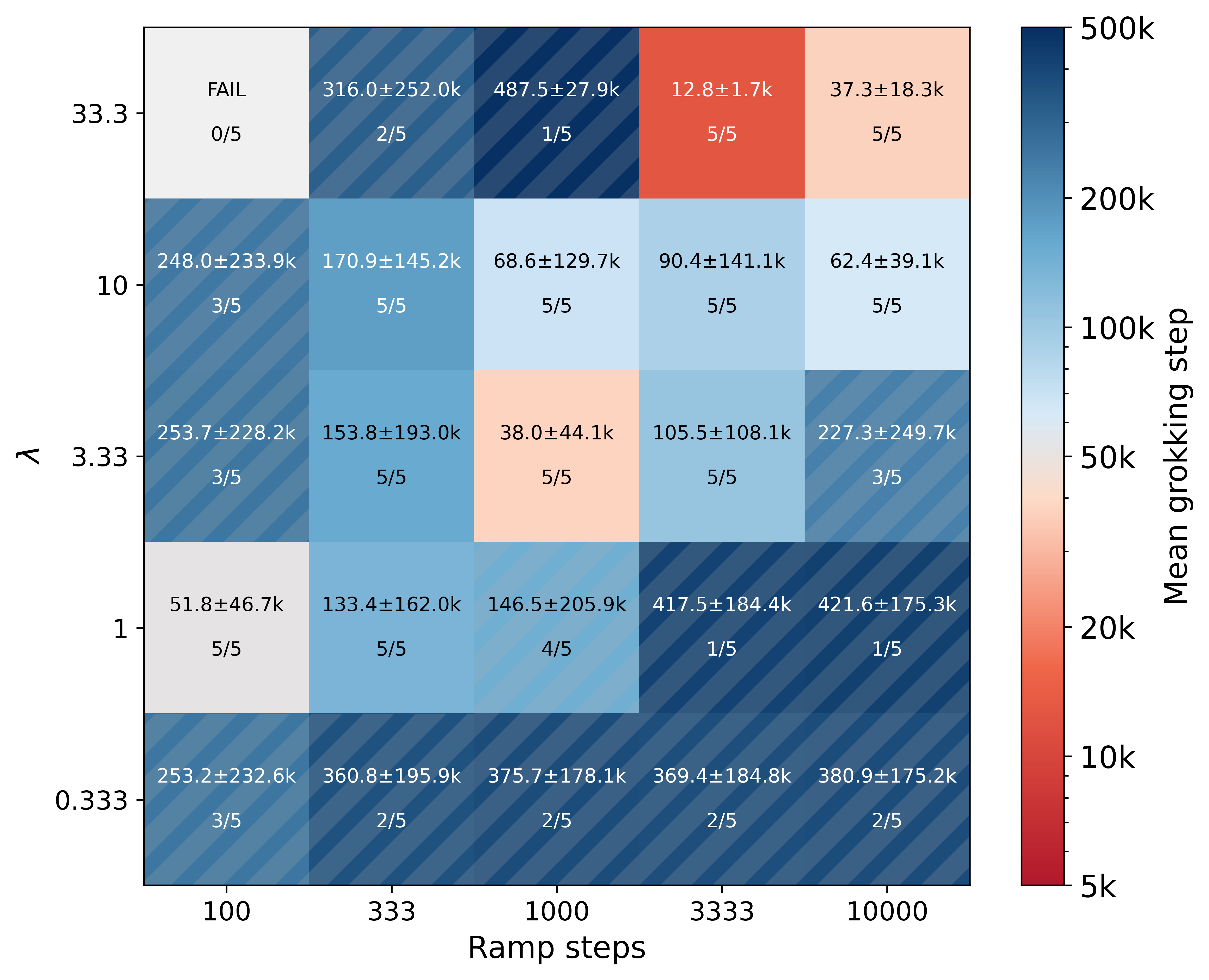}

\makebox[0.40\textwidth][c]{\small (a) Division: schedule sweep}
\hspace{0.005\textwidth}
\makebox[0.40\textwidth][c]{\small (b) Permutation composition: schedule sweep}


\includegraphics[width=0.40\textwidth]{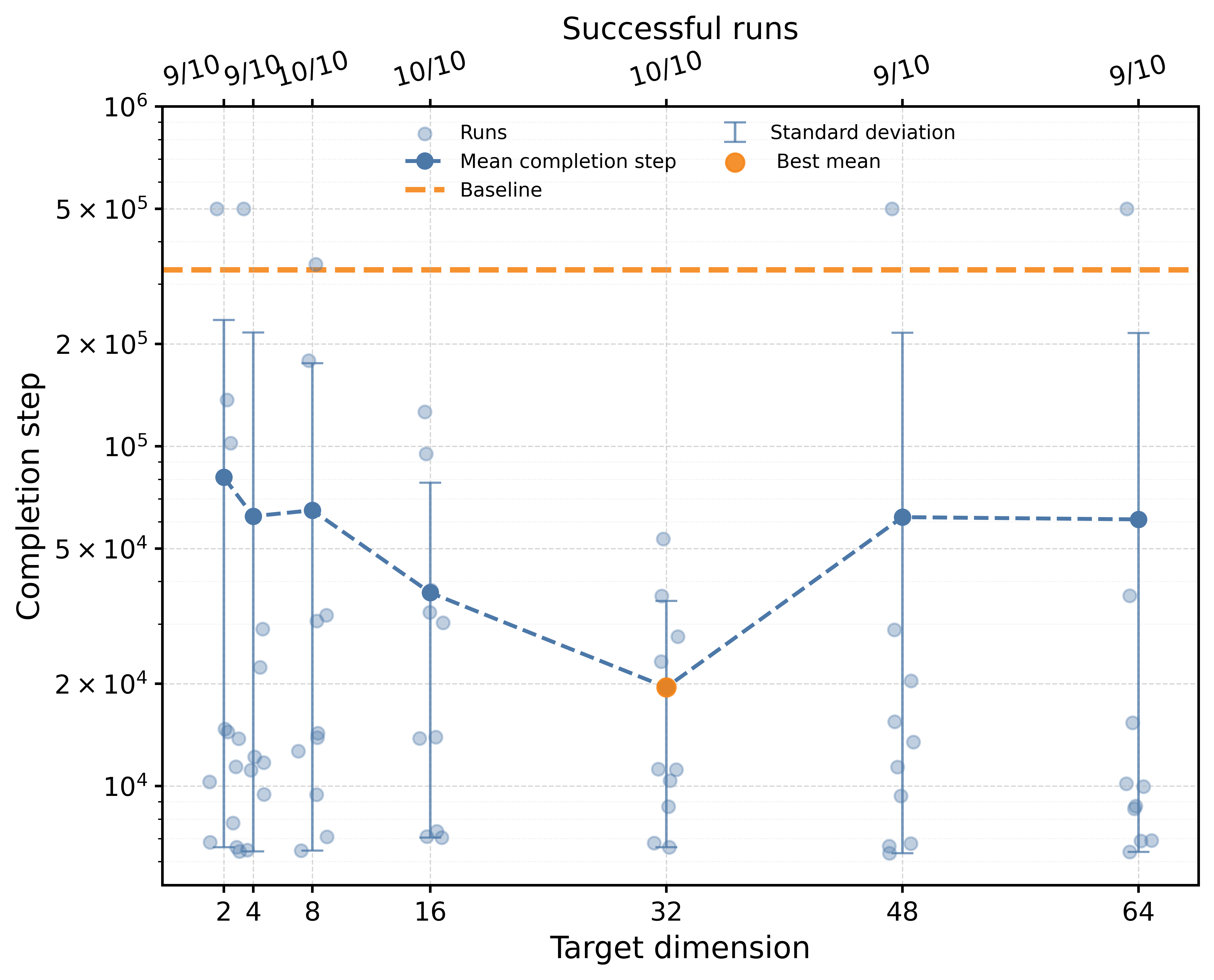}
\hspace{0.005\textwidth}
\includegraphics[width=0.40\textwidth]{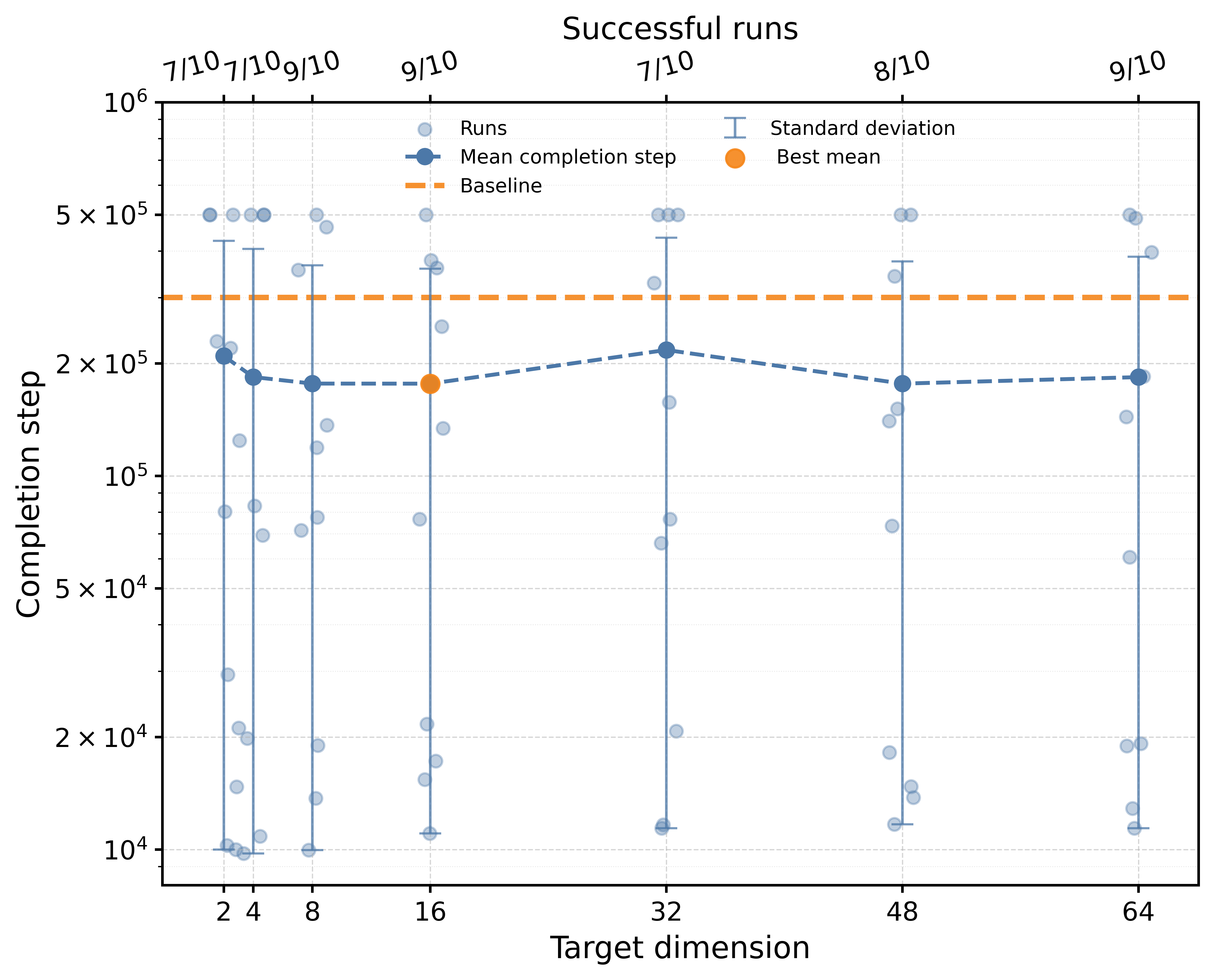}

\makebox[0.40\textwidth][c]{\small (c) Division: dimensionality sweep}
\hspace{0.005\textwidth}
\makebox[0.40\textwidth][c]{\small (d) Permutation composition: dimensionality sweep}


\includegraphics[width=0.40\textwidth]{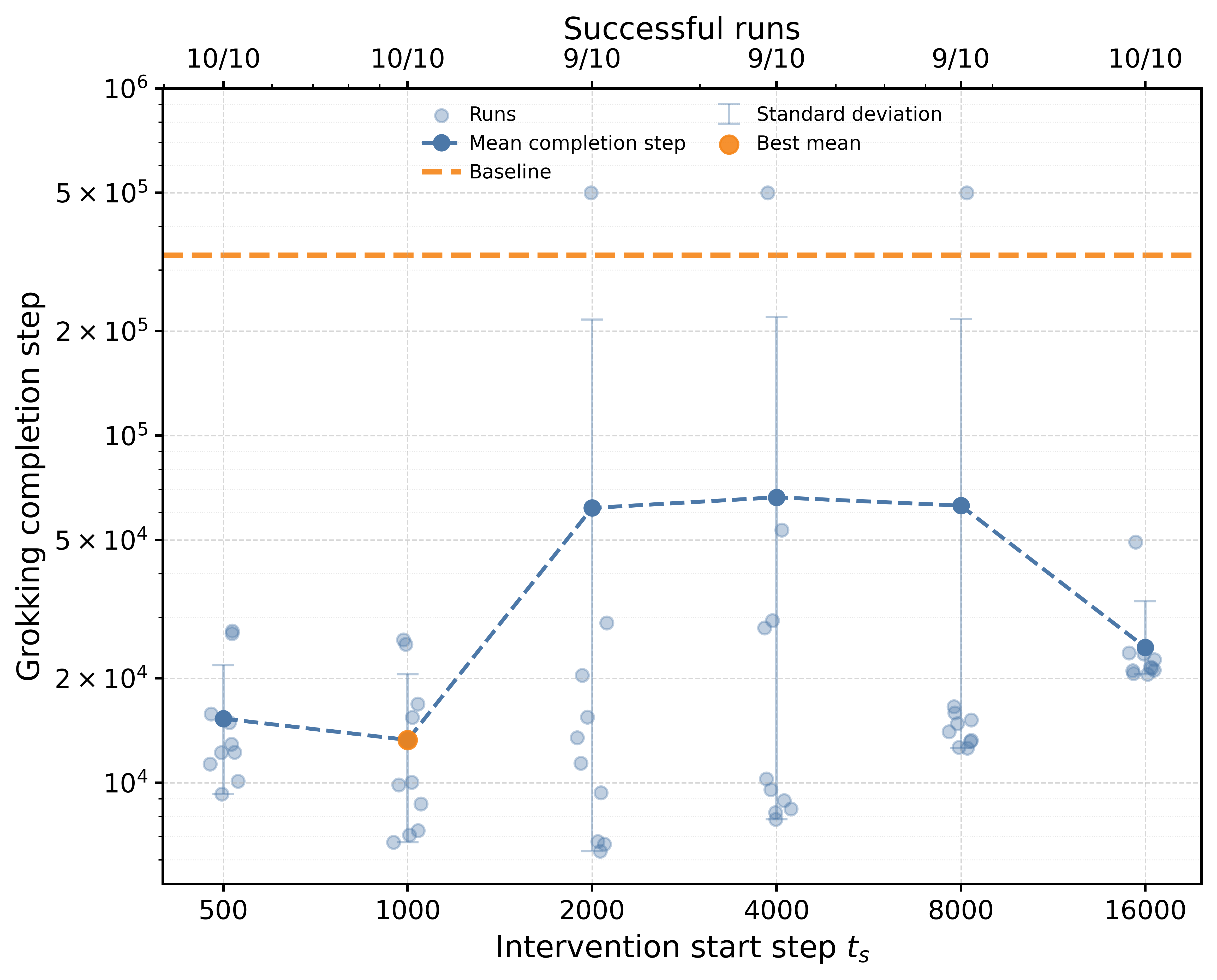}
\hspace{0.005\textwidth}
\includegraphics[width=0.40\textwidth]{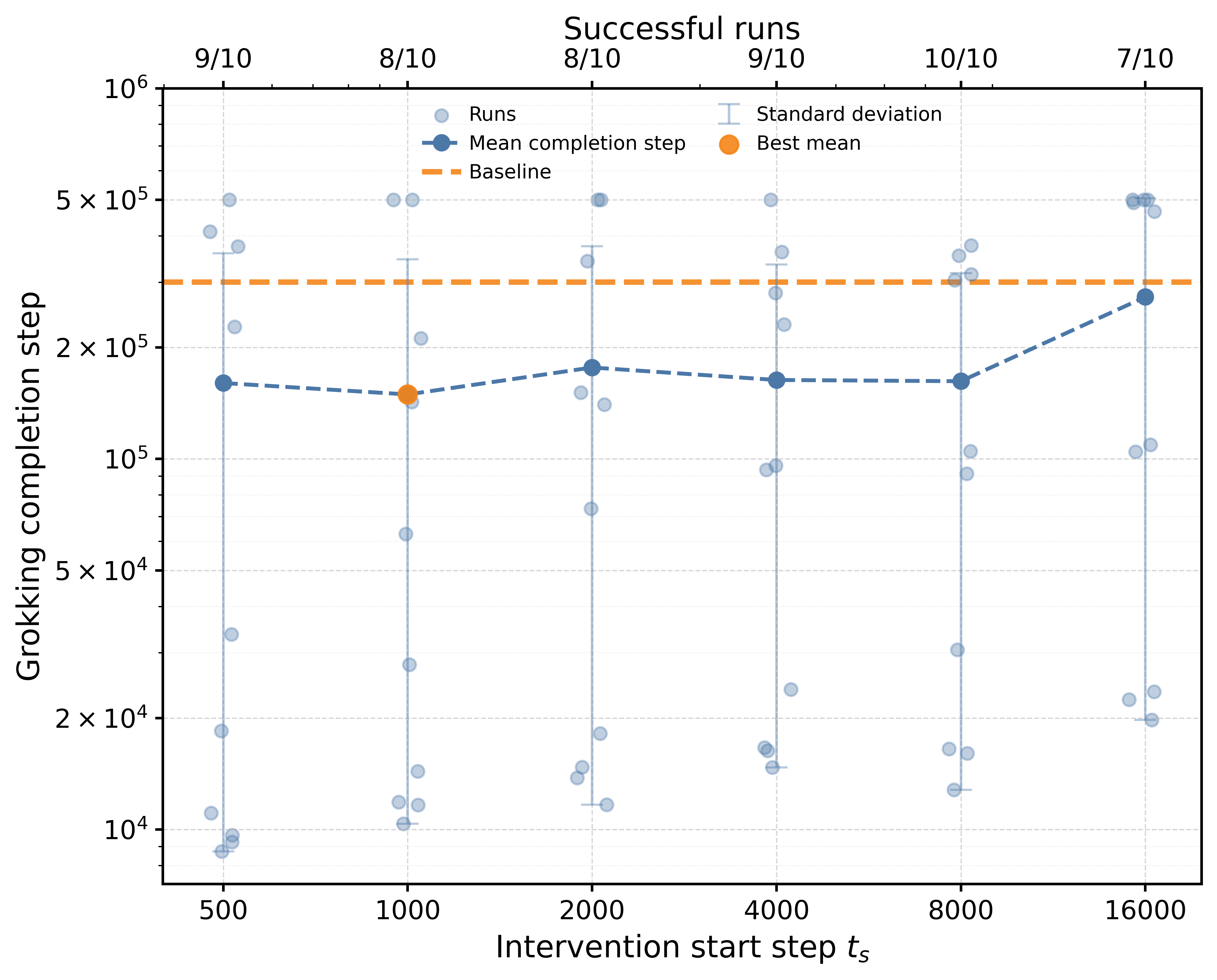}

\makebox[0.40\textwidth][c]{\small (e) Division: intervention timing}
\hspace{0.005\textwidth}
\makebox[0.40\textwidth][c]{\small (f) Permutation composition: intervention timing}

\caption{
Additional MLP experiments on modular division and permutation composition.
Top row: schedule sweeps over regularization strength $\lambda$ and ramp duration.
Middle row: target-dimensionality sweeps. 
Bottom row: intervention-start sweeps.
Consistent with the modular addition results reported in the main paper, Geometric Dimensionality Regularization (GeomDR) substantially accelerates grokking on both modular division and permutation composition. The effect remains robust across a broad range of schedules, target dimensionalities, and intervention times.
}
\label{fig:appendix_mlp_additional_tasks}

\end{figure*}

For the MLP architecture, we first evaluated the effect of the regularization schedule by varying the final regularization strength $\lambda_{\max}$ and ramp duration $T_{\mathrm{ramp}}$. We then performed target-dimensionality sweeps using a representative (not necessarily the best) schedule from the schedule-ablation study for each task and subsequently investigated the effect of intervention timing by varying the activation step $t_s$.

Table~\ref{tab:mlp_ablations} summarizes the effects of Geometric Dimensionality Regularization (GeomDR) across three grokking tasks using the MLP architecture. We evaluate two classes of interventions: target-dimensionality sweeps and intervention-time sweeps. Schedule-sweep results are omitted from the table because they are reported separately as heatmaps in Figure~\ref{fig:appendix_mlp_additional_tasks}. This figure additionally provides the corresponding dimensionality and intervention-timing visualizations for the modular division and permutation composition tasks.

For modular addition, schedule-sweep experiments fixed the target dimensionality at $d^*=16$ and the intervention-start step at $t_s=2000$, while varying the regularization strength $\lambda_{\max}\in\{0.333,1,3.33,10,33.3\}$ and ramp duration $T_{\mathrm{ramp}}\in\{100,333,1000,3333,10000\}$. Each configuration was evaluated across five random seeds $\{0,1,2,112,1122\}$. Dimensionality-sweep experiments used the representative schedule ($\lambda_{\max}=1$, $T_{\mathrm{ramp}}=333$, $t_s=2000$) and varied the target dimensionality over $d^*\in\{2,4,8,16,32,48,64\}$. Each configuration was evaluated across ten random seeds $\{0,1,2,3,4,5,6,7,8,9\}$. Intervention-timing experiments used the same schedule and target dimensionality ($\lambda_{\max}=1$, $T_{\mathrm{ramp}}=333$, $d^*=16$) while varying the intervention-start step over $t_s\in\{500,1000,2000,4000,8000,16000\}$, again using ten random seeds $\{0,1,2,3,4,5,6,7,8,9\}$.

For modular division, the schedule-ablation study followed the same protocol as modular addition, using $d^*=16$ and $t_s=2000$ while varying $\lambda_{\max}$ and $T_{\mathrm{ramp}}$ over the same ranges. Each configuration was evaluated across five random seeds $\{0,1,2,112,1122\}$. Dimensionality-sweep experiments used the representative schedule ($\lambda_{\max}=3.33$, $T_{\mathrm{ramp}}=1000$, $t_s=2000$) and varied the target dimensionality over $d^*\in\{2,4,8,16,32,48,64\}$. Each configuration was evaluated across ten random seeds $\{0,1,2,3,4,5,6,7,8,9\}$. Intervention-timing experiments used the same schedule and varied the intervention-start step over $t_s\in\{500,1000,2000,4000,8000,16000\}$, while fixing $\lambda_{\max}=3.33$, $T_{\mathrm{ramp}}=1000$, and $d^*=48$. Each configuration was evaluated across ten random seeds $\{0,1,2,3,4,5,6,7,8,9\}$.

For the permutation composition task, the schedule-ablation study followed the same protocol as modular addition, except that the target dimensionality was fixed at $d^*=48$ to reflect the larger hidden representation dimension. Dimensionality-sweep experiments used the representative schedule ($\lambda_{\max}=1$, $T_{\mathrm{ramp}}=333$, $t_s=2000$) and varied the target dimensionality over $d^*\in\{2,4,8,16,32,48,64\}$.  Intervention-timing experiments used the same schedule and target dimensionality ($\lambda_{\max}=1$, $T_{\mathrm{ramp}}=333$, $d^*=48$) while varying the intervention-start step over $t_s\in\{500,1000,2000,4000,8000,16000\}$. 

\begin{table*}[b!]
\centering

\caption{
Ablation studies of Geometric Dimensionality Regularization (GeomDR) on three grokking tasks using the Transformer architecture. The table reports dimension sweeps (varying the target dimensionality $d^*$), regularization-strength sweeps (varying the final GeomDR coefficient $\lambda_{\max}$), and intervention-time sweeps (varying the GeomDR activation step $t_s$). Values denote the mean grokking step $\pm$ one standard deviation (in thousands of optimization steps). Results are computed from 5 random seeds. The baseline corresponds to standard Transformer training without Geometric Dimensionality Regularization (GeomDR). Success/Total columns report the number of runs that satisfied the grokking criterion out of all runs. For runs that did not satisfy the criterion within the training budget, the maximum budget of 200K optimization steps was used when computing the reported mean and standard deviation.
}

\label{tab:transformer_ablations}

\begin{tabular}{lcc cc cc}
\toprule
&
\multicolumn{2}{c}{Addition}
&
\multicolumn{2}{c}{Division}
&
\multicolumn{2}{c}{Permutation}
\\

\cmidrule(lr){2-3}
\cmidrule(lr){4-5}
\cmidrule(lr){6-7}

Setting
&
Step (K)
&
Success/Total
&
Step (K)
&
Success/Total
&
Step (K)
&
Success/Total
\\

\midrule

Baseline
&
$66.4 \pm 12.0$
&
$5/5$
&
$89.4 \pm 27.3$
&
$5/5$
&
$72.6 \pm 33.0$
&
$5/5$
\\

\midrule
\multicolumn{7}{c}{\textbf{Dimension Sweep}}
\\
\midrule

2
&
$41.2 \pm 19.2$
&
$5/5$
&
$59.9 \pm 14.1$
&
$5/5$
&
$85.2 \pm 58.0$
&
$5/5$
\\

4
&
$\mathbf{36.7 \pm 9.1}$
&
$5/5$
&
$\mathbf{58.8 \pm 27.9}$
&
$5/5$
&
$155.1 \pm 174.4$
&
$4/5$
\\

8
&
$46.0 \pm 9.8$
&
$5/5$
&
$71.5 \pm 39.5$
&
$5/5$
&
$131.6 \pm 184.2$
&
$4/5$
\\

16
&
$136.7 \pm 181.7$
&
$4/5$
&
$70.8 \pm 30.2$
&
$5/5$
&
$\mathbf{68.0 \pm 29.5}$
&
$5/5$
\\

\midrule
\multicolumn{7}{c}{\textbf{Lambda Sweep}}
\\
\midrule

33.3
&
$74.0 \pm 63.0$
&
$4/5$
&
$\mathbf{49.1 \pm 13.4}$
&
$5/5$
&
$120.6 \pm 59.9$
&
$4/5$
\\

10
&
$34.6 \pm 6.2$
&
$5/5$
&
$67.8 \pm 51.7$
&
$5/5$
&
$74.8 \pm 63.5$
&
$4/5$
\\

3.33
&
$\mathbf{34.5 \pm 7.4}$
&
$5/5$
&
$88.4 \pm 59.6$
&
$4/5$
&
$97.3 \pm 66.2$
&
$4/5$
\\

1
&
$38.2 \pm 15.4$
&
$5/5$
&
$65.5 \pm 28.3$
&
$5/5$
&
$\mathbf{48.9 \pm 9.1}$
&
$\mathbf{5/5}$
\\

0.333
&
$35.9 \pm 1.9$
&
$5/5$
&
$52.9 \pm 9.6$
&
$5/5$
&
$79.5 \pm 61.2$
&
$4/5$
\\
\midrule
\multicolumn{7}{c}{\textbf{Intervention Time Sweep}}
\\
\midrule

500
&
$47.2 \pm 13.0$
&
$5/5$
&
$112.0 \pm 66.1$
&
$4/5$
&
$71.3 \pm 65.2$
&
$4/6$
\\

1000
&
$\mathbf{33.0 \pm 2.5}$
&
$5/5$
&
$101.1 \pm 54.0$
&
$4/5$
&
$93.6 \pm 73.0$
&
$4/5$
\\

2000
&
$41.1 \pm 11.7$
&
$5/5$
&
$63.8 \pm 23.3$
&
$5/5$
&
$80.2 \pm 26.4$
&
$5/5$
\\

4000
&
$35.2 \pm 3.3$
&
$5/5$
&
$77.4 \pm 61.4$
&
$4/5$
&
$77.1 \pm 52.8$
&
$5/5$
\\

8000
&
$36.4 \pm 8.2$
&
$5/5$
&
$\mathbf{48.7 \pm 8.8}$
&
$5/5$
&
$102.6 \pm 54.4$
&
$4/5$
\\

16000
&
$78.7 \pm 60.9$
&
$4/5$
&
$59.0 \pm 21.0$
&
$5/5$
&
$\mathbf{65.1 \pm 14.2}$
&
$5/5$
\\

\bottomrule
\end{tabular}
\end{table*}
Across all tasks, GeomDR substantially accelerates grokking relative to the baseline. For modular addition, the baseline model reaches the grokking criterion after $362.1 \pm 111.0$K optimization steps (4/5 successful runs), whereas multiple GeomDR configurations reduce the grokking time to $7.0 \pm 1.3$K steps (5/5 successful runs). 

The schedule-ablation results exhibit a relatively smooth dependence on both the regularization strength $\lambda_{\max}$ and ramp duration $T_{\mathrm{ramp}}$. Rather than displaying highly irregular behavior, performance changes gradually across neighboring configurations, indicating that GeomDR is not overly sensitive to precise hyperparameter choices. Across all tasks, successful configurations form contiguous regions in the hyperparameter space, suggesting a broad basin of effective schedules.

A consistent trend is that more challenging tasks require stronger regularization. For modular addition, optimal performance is achieved with relatively weak regularization ($\lambda_{\max}=1$), whereas modular division benefits from intermediate regularization strengths ($\lambda_{\max}=3.33$). The permutation composition task, which exhibits the most difficult grokking dynamics among the considered MLP benchmarks, achieves its best performance with substantially stronger regularization ($\lambda_{\max}=33.3$). These results suggest that the amount of geometric compression required to accelerate grokking increases with task complexity.

The dimensionality sweep reveals that GeomDR is effective across a broad range of target dimensions. For modular addition, the fastest and most stable performance is obtained for $d^* \in \{16,32\}$. For modular division, the best performance is achieved at $d^*=32$, although substantial improvements are observed across nearly the entire range of tested target dimensionalities.  For permutation composition, performance is strongest at an intermediate target dimensionality $d^*=16$, while both excessively strong and excessively weak compression lead to slower grokking and reduced success rates. Extremely small target dimensions may impose excessive representational constraints, while excessively large target dimensions may provide insufficient geometric pressure. These results indicate that the benefits of GeomDR are not restricted to a narrowly tuned target dimensionality and that effective geometric regularization can be achieved across a broad range of target dimensions.

The intervention-time sweep demonstrates that GeomDR remains effective across a broad range of activation times. For modular addition, the best performance is achieved at $t_s=2000$, with similarly strong results at $t_s=4000$.  For modular division, the best mean performance obtained at
$t_{s}=1000$. For permutation composition, the lowest mean grokking step is observed at $t_s=1000$, while the most reliable performance is obtained at $t_s=8000$, which achieves a $10/10$ success rate. Overall, all intervention times substantially outperform the baseline. Across the three tasks, early-to-intermediate interventions generally provide the largest acceleration. One possible explanation is that prolonged memorization makes the learned representations increasingly specialized and less amenable to subsequent geometric restructuring. GeomDR may therefore be most effective after useful task structure has begun to emerge, but before the representation geometry becomes comparatively rigid.

\subsection{Transformer}

\label{app:transformer_ablations}
For the Transformer architecture, we adopted a different tuning procedure from that used for the MLP. To reduce the size of the hyperparameter search space, we fixed the ramp duration at $T_{\mathrm{ramp}}=1000$ throughout all Transformer experiments. We first performed target-dimensionality sweeps to identify the best-performing target dimensionality $d^*$. Using this best configuration, we then optimized the regularization strength $\lambda_{\max}$. Finally, using the best-performing dimensionality and regularization strength, we evaluated the effect of intervention timing by varying the activation step $t_s$.

Table~\ref{tab:transformer_ablations} summarizes ablation studies of Geometric Dimensionality Regularization (GeomDR) on the Transformer architecture across the modular addition, modular division, and permutation composition tasks.

For Transformer-based modular addition, dimensionality-sweep experiments fixed the intervention-start step at $t_s=1000$, the ramp duration at $T_{\mathrm{ramp}}=1000$, and the regularization strength at $\lambda_{\max}=1$, while varying the target dimensionality over $d^*\in\{2,4,8,16\}$.  Using the best-performing dimensionality ($d^*=4$), we then performed a regularization-strength sweep over $\lambda_{\max}\in\{0.333,1,3.33,10,33.3\}$ while keeping $t_s=1000$ fixed. Finally, intervention-timing experiments used the best-performing configuration ($d^*=4$, $\lambda_{\max}=3.33$) and varied the activation step over $t_s\in\{500,1000,2000,4000,8000,16000\}$. Each configuration was evaluated across five random seeds $\{0,1,2,112,1122\}$.

For Transformer-based modular division, we followed the same experimental procedure as for modular addition. Dimensionality-sweep experiments varied the target dimensionality over $d^*\in\{2,4,8,16\}$ while fixing $t_s=1000$,  and $\lambda_{\max}=1$. Using the best-performing dimensionality ($d^*=4$), we then performed a regularization-strength sweep over $\lambda_{\max}\in\{0.333,1,3.33,10,33.3\}$. Finally, intervention-timing experiments varied the activation step over $t_s\in\{500,1000,2000,4000,8000,16000\}$ using the best-performing configuration from the previous stage ($d^*=4$, $\lambda_{\max}=33.3$). Each configuration was evaluated across five random seeds $\{0,1,2,112,1122\}$.

For Transformer-based permutation composition, we followed the same experimental procedure as for modular addition. Dimensionality-sweep experiments varied the target dimensionality over $d^*\in\{2,4,8,16\}$ while fixing $t_s=1000$, and $\lambda_{\max}=1$. Using the best-performing dimensionality ($d^*=16$), we then performed a regularization-strength sweep over $\lambda_{\max}\in\{0.333,1,3.33,10,33.3\}$. Finally, intervention-timing experiments varied the activation step over $t_s\in\{500,1000,2000,4000,8000,16000\}$ using the best-performing configuration from the previous stage ($d^*=16$, $\lambda_{\max}=1$). Each configuration was evaluated across five random seeds $\{0,1,2,112,1122\}$.

Across all three tasks, GeomDR consistently accelerates grokking relative to the baseline Transformer. For modular addition, the baseline model requires $66.4 \pm 12.0$K optimization steps to satisfy the grokking criterion, whereas the best GeomDR configuration reduces this value to $33.0 \pm 2.5$K optimization steps, corresponding to roughly a twofold acceleration. Similar improvements are observed for modular division, where the grokking step decreases from $89.4 \pm 27.3$K to $48.7 \pm 8.8$K optimization steps. For permutation composition, the strongest GeomDR configurations reduce the grokking step from $72.6 \pm 33.0$K to $48.9 \pm 9.1$K optimization steps. Although the improvements are generally smaller than those observed for the MLP architecture, these results demonstrate that the benefits of geometric regularization extend beyond modular arithmetic and remain effective on structured combinatorial problems.

The dimensionality sweep reveals that moderate target dimensionalities provide the strongest improvements for the arithmetic tasks. For both modular addition and modular division, target dimensions $d^*\in\{2,4\}$ yield the fastest mean grokking times, while larger dimensions remain beneficial but generally produce weaker acceleration. In contrast, permutation composition achieves its best performance at the larger target dimensionality $d^*=16$. This difference suggests that more complex tasks may benefit from retaining a higher-dimensional representation space, whereas simpler arithmetic tasks can be effectively accelerated through stronger geometric compression. 

For modular addition, the fastest mean grokking time is obtained at $\lambda_{\max}=3.33$, whereas modular division achieves its best performance at $\lambda_{\max}=33.3$. In contrast, permutation composition performs best at the more moderate value $\lambda_{\max}=1$.  These results further suggest that the optimal strength of geometric regularization is task-dependent and may vary with the complexity and structure of the underlying problem.

The intervention-time sweep shows that the timing of the geometric intervention also influences performance. For modular addition, activating GeomDR after an initial unconstrained training phase ($t_s=1000$) produces the fastest and most consistent results. For modular division, the lowest mean grokking step obtained at $t_s=8000$. For permutation composition, the strongest performance is achieved at $t_s=16000$, although several intervention times produce broadly similar results. Taken together, these findings suggest that the optimal intervention time is task-dependent and may reflect differences in the duration of the memorization phase preceding generalization. 

Compared with the MLP results summarized in Table~\ref{tab:mlp_ablations}, the gains obtained with GeomDR are more modest for the Transformer architecture. Whereas GeomDR often accelerates grokking by more than an order of magnitude in MLPs, he Transformer typically exhibits improvements ranging from roughly $1.5$ times to $2$times. Nevertheless, the qualitative trends remain consistent across architectures, with geometric interventions reliably reducing grokking time across all three tasks.

One possible explanation is that Transformers already possess strong inductive biases toward structured representations through self-attention, residual connections, and normalization layers.  As a result, the baseline Transformer groks substantially faster than the corresponding MLP, leaving less room for geometric regularization to further accelerate the memorization-to-generalization transition. While this interpretation remains speculative, it suggests that the effectiveness of GeomDR may depend in part on the extent to which the underlying architecture already promotes compressed and task-relevant representations.

\subsection{Width Ablation}

To evaluate the dependence of GeomDR on model capacity, we performed a width ablation study on the modular addition task for both MLP and Transformer architectures. For MLPs, width refers to the hidden dimension of the residual network, while for Transformers it refers to the model dimension $d_{\mathrm{model}}$. We evaluated MLP widths $\{32,64,128,256\}$ and Transformer widths $\{16,32,64,128\}$ while keeping all other architectural components fixed. For MLPs, we used the representative configuration identified in the schedule sweep ($d^*=16$, $\lambda_{\max}=1$, $T_{\mathrm{ramp}}=333$). For Transformers, we used the best-performing configuration identified during Transformer tuning, corresponding to a target dimensionality of $d^*=4$, regularization strength $\lambda_{\max}=3.33$, intervention-start step $t_s=1000$, and ramp duration $T_{\mathrm{ramp}}=1000$. Each experiment was repeated with five random seeds $\{0,1,2,112,1122\}$.

Table~\ref{tab:width_ablation} summarizes the results. For MLPs, GeomDR consistently accelerates grokking across all tested widths. At width $32$, the baseline failed to satisfy the grokking criterion in any run within the training budget, whereas GeomDR achieved a mean grokking step of $21.6$K with a $5/5$ success rate. A similar effect is observed at width $64$, where the baseline again failed in all runs while GeomDR achieved a mean grokking step of $19.3$K and successfully grokked in all runs. At width $128$, GeomDR reduces the mean grokking step from $333.0$K to $6.8$K while increasing the success rate from $4/5$ to $5/5$. At width $256$, the mean grokking step decreases from $253.8$K to $10.4$K, again achieving a $5/5$ success rate. Overall, the intervention substantially accelerates grokking and improves reliability across a broad range of MLP capacities.

The Transformer results exhibit a markedly different pattern. At width $32$, which corresponds to the default Transformer configuration used throughout the main experiments, GeomDR substantially accelerates grokking, reducing the mean grokking step from $65.4$K to $34.1$K while maintaining a $5/5$ success rate. However, this benefit does not persist at larger widths. At width $64$, the intervention succeeds in only $2/5$ runs, yielding a mean grokking step of $139.9$K when unsuccessful runs are included in the summary statistics. At width $128$, only a single run satisfies the grokking criterion, resulting in an average grokking step of $174.6$K. In both cases, performance is substantially worse than the corresponding baseline models. The narrowest Transformer configuration (width $16$) fails to grok within the training budget both with and without intervention.

Overall, the width ablation shows that GeomDR is highly robust across MLP capacities, suggesting that the intervention acts on a geometric property that remains stable under changes in representational width. Transformer results reveal a more architecture-dependent behavior. While GeomDR can accelerate grokking in Transformers under suitable settings, the fixed low dimensional target used here does not scale reliably with model width and can even degrade performance for larger models. These findings indicate that representation geometry remains a useful control signal beyond MLPs, but that effective geometric interventions are likely architecture-dependent. These findings indicate that representation geometry remains a useful control signal beyond MLPs, but that effective geometric interventions are likely architecture-dependent and may require retuning as model width increases.

\begin{table}[t]
\centering
\caption{
Width ablation on modular addition. Results are reported as mean grokking step $\pm$ standard deviation across five random seeds. Succ/Tot denotes the number of runs that satisfied the grokking criterion.
}
\label{tab:width_ablation}

\small

\textbf{(a) MLP}

\vspace{1mm}

\begin{tabular}{lcccc}
\toprule
&
\multicolumn{2}{c}{Baseline}
&
\multicolumn{2}{c}{GeomDR ($d^*=16$)}
\\
\cmidrule(lr){2-3}
\cmidrule(lr){4-5}
Width &
Step (K) &
Succ/Tot &
Step (K) &
Succ/Tot \\
\midrule
32  & $500.0 \pm 0.0$   & $0/5$ & $21.6 \pm 5.3$   & $5/5$ \\
64  & $500.0 \pm 0.0$   & $0/5$ & $19.3 \pm 23.5$  & $5/5$ \\
128& $333.0 \pm 139.2$ & $4/5$ & $6.8 \pm 0.8$    & $5/5$ \\
256 & $253.8 \pm 146.4$ & $4/5$ & $10.4 \pm 5.6$   & $5/5$ \\
\bottomrule
\end{tabular}

\vspace{3mm}

\textbf{(b) Transformer}

\vspace{1mm}

\begin{tabular}{lcccc}
\toprule
&
\multicolumn{2}{c}{Baseline}
&
\multicolumn{2}{c}{GeomDR ($d^*=4$)}
\\
\cmidrule(lr){2-3}
\cmidrule(lr){4-5}
Width &
Step (K) &
Succ/Tot &
Step (K) &
Succ/Tot \\
\midrule
16  & $200.0 \pm 0.0$   & $0/5$ & $200.0 \pm 0.0$   & $0/5$ \\
32 & $65.4 \pm 7.2$    & $5/5$ & $34.1 \pm 5.7$    & $5/5$ \\
64  & $43.7 \pm 5.8$    & $5/5$ & $139.9 \pm 83.7$  & $2/5$ \\
128 & $58.6 \pm 26.2$   & $5/5$ & $174.6 \pm 56.9$  & $1/5$ \\
\bottomrule
\end{tabular}

\end{table}

\subsection{Collapse--Grokking Lag Analysis}
\label{app:collapse_grokking_lag}
Table~\ref{tab:collapse_grok} reports the quantitative analysis of the temporal separation between representation collapse and grokking. Collapse time is defined as the first step at which the effective dimensionality of the deepest hidden layer ($L_4$) decreases below 50\% of its initial value. The initial dimensionality is computed from the first recorded representation snapshot at initialization (step 0) using the effective dimension metric. The lag is defined as
$t_{\mathrm{grok}}-t_{\mathrm{collapse}}$.
All experiments use the modular addition task and the MLP architecture described earlier. Values are averaged over five random seeds in
$\{0,1,2,112,1122\}$.
For this setting, baseline models exhibit a large delay between dimensionality collapse and successful generalization, whereas GeomDR substantially reduces this temporal gap, bringing representation compression and generalization into closer alignment.
\begin{table}[!h]
\centering

\caption{
Timing of dimensionality collapse and grokking.
Collapse is defined as the first step at which the effective dimension
of the final hidden layer falls below 50\% of its initial value.
Lag denotes the temporal separation
$t_{\mathrm{grok}}-t_{\mathrm{collapse}}$.
All values are reported in thousands of training steps (K) as
mean $\pm$ standard deviation.
Succ/Tot denotes the number of runs satisfying the grokking criterion
out of the total number of runs.
For the baseline, lag statistics were computed over the four seeds
that successfully grokked; one non-grokking run was omitted from
the lag analysis.
}
\label{tab:collapse_grok}

\footnotesize
\setlength{\tabcolsep}{3pt}

\begin{tabular}{lcccc}
\toprule
Setting & Succ/Tot & Collapse (K) & Grok (K) & Lag (K) \\
\midrule

Baseline &
4/5 &
$19.1 \pm 2.0$ &
$283.4 \pm 33.5$ &
$264.3 \pm 33.8$
\\

\midrule

$d^*=64$ &
5/5 &
$3.7 \pm 0.3$ &
$\mathbf{5.9 \pm 1.1}$ &
$\mathbf{2.2 \pm 0.8}$
\\

$d^*=32$ &
5/5 &
$3.7 \pm 0.3$ &
$6.6 \pm 1.5$ &
$2.9 \pm 1.4$
\\

$d^*=16$ &
5/5 &
$3.9 \pm 0.4$ &
$6.9 \pm 1.6$ &
$3.0 \pm 1.4$
\\

$d^*=8$ &
5/5 &
$3.8 \pm 0.2$ &
$6.7 \pm 1.2$ &
$2.9 \pm 1.1$
\\

\bottomrule
\end{tabular}

\end{table}

\noindent
\begin{table*}[!t]
\centering
\caption{
Random-sampling ablation for Geometric Dimensionality Regularization (GeomDR) on modular addition using the MLP architecture. The covariance spectrum is estimated from a randomly sampled subset of hidden representations. Values report mean $\pm$ standard deviation.
}
\label{tab:scalability}
\begin{tabular}{lcccc}
\toprule
Method & Sample Size (\% of Data) & Success / Total & Runtime (s) & Grokking Step (K) \\
\midrule

Baseline &
- &
4/5 &
$12952.0 \pm 3154.4$ &
$362.1 \pm 111.0$ \\

GeomDR (100\%) &
2822 (100\%) &
5/5 &
$666.0 \pm 368.0$ &
$7.06 \pm 1.74$ \\

GeomDR (50\%) &
1411 (50\%) &
5/5 &
$447.2 \pm 274.9$ &
$7.36 \pm 3.72$ \\

GeomDR (25\%) &
705 (25\%) &
5/5 &
$276.7 \pm 50.6$ &
$\mathbf{6.34 \pm 1.05}$ \\

GeomDR (10\%) &
282 (10\%) &
5/5 &
$259.6 \pm 94.3$ &
$6.81 \pm 1.96$ \\

GeomDR (5\%) &
141 (5\%) &
5/5 &
$\mathbf{247.5 \pm 76.9}$ &
$7.89 \pm 1.87$ \\

GeomDR (1\%) &
28 (1\%) &
5/5 &
$1088.7 \pm 366.8$ &
$35.76 \pm 9.65$ \\

\bottomrule
\end{tabular}
\end{table*}
\FloatBarrier
\section{Scalability}
\label{app:scalability}

For large datasets, computing covariance spectra using all available representations may become prohibitively expensive. We therefore investigate whether GeomDR remains effective when the covariance spectrum is estimated from a randomly sampled subset of representations.

Indeed, the computational complexity of GeomDR for a single layer is dominated by covariance estimation and eigendecomposition. Given $N$ representations of dimension $d$, computing the covariance matrix requires $O(Nd^2)$ operations, while eigendecomposition of the resulting $d \times d$ covariance matrix requires $O(d^3)$ operations. The overall complexity is therefore

\[
O(Nd^2 + d^3).
\]

In the regimes considered in this work, the number of representations is substantially larger than the representation dimension ($N \gg d$), so the covariance computation dominates and the effective complexity scales approximately as $O(Nd^2)$. If the covariance spectrum is estimated using only a fraction $r$ of the available representations, the complexity for one layer becomes

\[
O(rNd^2 + d^3),
\]

implying an approximately linear reduction in computational cost with respect to the sampling ratio $r$.

Experiments are performed on the modular addition task using the MLP architecture. The intervention uses $\lambda=3.33$, target dimensionality $d^*=16$, activation step $t_s=2000$, and $T_{ramp} = 1000$. At each optimization step, a fraction

\[
r \in \{1\%,5\%,10\%,25\%,50\%,100\%\}
\]
of the training representations is sampled uniformly without replacement, and the GeomDR penalty is computed using only the sampled subset. For the modular addition task, the training set contains $2822$ examples, corresponding to sample sizes ranging from $28$ representations at $1\%$ sampling to the full training set at $100\%$ sampling.
Table~\ref{tab:scalability} reports both wall-clock runtime and grokking step. Across a broad range of sampling ratios, GeomDR remains highly effective despite using substantially fewer representations for covariance estimation. Sampling ratios between $5\%$ and $50\%$ yield grokking times comparable to those obtained with full-spectrum estimation, with successful generalization typically occurring after approximately $6$--$8$K optimization steps. The fastest average grokking time is observed at $25\%$ sampling, reaching the grokking criterion after $6.34 \pm 1.05$K steps, compared to $7.06 \pm 1.74$K steps for full-spectrum estimation. However, the differences in grokking steps among sampling ratios in the $5\%$--$50\%$ range are relatively small, indicating that accurate covariance estimation can be achieved from substantially reduced representation subsets without materially affecting grokking acceleration.

Subsampling also substantially reduces computational cost. While full-spectrum GeomDR requires $666.0 \pm 368.0$ seconds of wall-clock time (one CPU), sampling ratios between $5\%$ and $25\%$ reduce runtime to approximately $250$--$280$ seconds, representing a $2.4\times$--$2.7\times$ speedup relative to full-spectrum GeomDR. Compared with the baseline model without GeomDR, these configurations reduce runtime by roughly a factor of $50$. The 1\% condition exhibits substantially higher runtime due to
its much later grokking time, requiring considerably more
optimization steps before satisfying the grokking criterion.
These results suggest that the geometric signal exploited by GeomDR can be estimated accurately from relatively small random subsets of representations. In practice, covariance estimation using only $5\%$--$25\%$ of available representations is sufficient to recover nearly the full benefit of GeomDR while substantially reducing computational overhead. This observation indicates that precise estimation of the full covariance spectrum is not required for effective geometric control, improving the practical scalability and computational efficiency
of the method. 

These findings suggest that GeomDR may be compatible with standard mini-batch optimization. The random sampling results show that only a small subset of representations is required to estimate the geometric signal driving the regularizer. Although this hypothesis was not evaluated directly in the present work, each mini-batch can be viewed as a random sample of the representation distribution, suggesting that the leading covariance structure may be estimated sufficiently accurately from batch-level statistics. If so, GeomDR could be implemented efficiently in large-scale settings without computing covariance matrices over the entire dataset.

\section{Comparison with Grokking Acceleration Methods}
\label{app:comparison_grokfast}

We compare Geometric Dimensionality Regularization (GeomDR) with two commonly studied grokking interventions: weight decay and GrokFast \cite{lee2024grokfast}. For weight decay, we evaluate $\mathrm{WD}\in\{0.01,0.03,0.1,0.3,1.0\}$ and report the best-performing configuration ($\mathrm{WD}=1.0$). For GrokFast, we evaluate amplification coefficients
$\lambda \in \{0.05,0.1,0.5,1.0,5.0\}$
using the default exponential moving average coefficient
$\alpha=0.999$ and baseline weight decay $\mathrm{WD}=0.1$, and report the best-performing configuration identified in the sweep ($\lambda=1.0$). GeomDR uses the best-performing modular-addition configuration identified in the ablation studies ($d^*=16$, $t_s=2000$, $T_{\mathrm{ramp}}=1000$, $\lambda_{\max}=1.0$).

Results are summarized in Table~\ref{tab:comparison_methods}. The best weight-decay configuration substantially accelerates grokking relative to the baseline, reaching the grokking criterion after $27.2 \pm 43.9$K optimization steps. GrokFast also improves performance relative to the baseline, achieving a mean grokking step of $372.1 \pm 37.6$K and successfully grokking in all runs. However, the magnitude of this improvement is considerably smaller than that obtained with weight decay or GeomDR.

\begin{table}[t]
\centering

\caption{
Comparison of methods for accelerating grokking on modular addition using the MLP architecture.
Values denote mean grokking step $\pm$ standard deviation (thousands of optimization steps).
Succ/Tot indicates the number of runs satisfying the grokking criterion.
For unsuccessful runs, the training budget of 500K steps was used when computing summary statistics.
}
\label{tab:comparison_methods}

\footnotesize
\setlength{\tabcolsep}{4pt}

\begin{tabular}{lccc}
\toprule
Method &
Best Params &
Step (K) &
Succ/Tot \\
\midrule

WD = 0
&
--
&
500.0 $\pm$ 0.0
&
0/5
\\

\midrule

Weight Decay
&
WD = 1.0
&
27.2 $\pm$ 43.9
&
5/5
\\

GrokFast
&
$\lambda = 1.0$
&
372.1 $\pm$ 37.6
&
5/5
\\

GeomDR
&
$\lambda_{\max}=1$
&
\textbf{7.0 $\pm$ 1.5}
&
\textbf{10/10}
\\

GrokFast + GeomDR
&
$\lambda=1.0$
&
500.0 $\pm$ 0.0
&
0/5
\\

\bottomrule
\end{tabular}

\end{table}
GeomDR provides the strongest acceleration, reaching the grokking criterion after $7.0 \pm 1.5$K optimization steps while maintaining a perfect success rate across all runs. We additionally evaluated a combined GrokFast+GeomDR intervention. In contrast to the individual methods, the combined configuration failed to satisfy the grokking criterion in any run within the available training budget, suggesting that the two interventions may interact unfavorably and over-constrain the optimization dynamics.

Increasing the weight-decay coefficient from $0.1$ to $1.0$ in the presence of GeomDR did not provide additional improvement, suggesting diminishing returns from combining stronger parameter-space regularization with explicit geometric regularization. These results indicate that direct control of representation geometry can provide a highly effective mechanism for accelerating grokking, substantially outperforming both tuned weight decay and GrokFast in the experimental setting considered here.

\clearpage
\section{Use of Large Language Models}

Large language models (LLMs) were used as writing and editing assistants during the preparation of this manuscript. Specifically, LLMs were used to improve the grammar, clarity, and presentation of the text.

The scientific content of this work, including the research questions, hypotheses, experimental design, implementation, data analysis, interpretation of results, and conclusions, was developed and verified by the author.

All AI-assisted text was reviewed and edited by the author prior to submission. The author assumes full responsibility for the accuracy and content of this manuscript.

\section{Code Availability}

Code is available at \url{https://github.com/maksimkazanskii/grokking}.

\end{document}